\documentclass[10pt,twocolumn,letterpaper]{article}

\usepackage{cvpr}
\usepackage{times}
\usepackage{epsfig}
\usepackage{graphicx}
\usepackage{amsmath}
\usepackage{amssymb}
\usepackage{multirow}
\usepackage{color}
\usepackage[toc,page]{appendix}

\usepackage[pagebackref=true,breaklinks=true,letterpaper=true,colorlinks,bookmarks=false]{hyperref}

\cvprfinalcopy 

\def\cvprPaperID{18} 

\definecolor{purple}{RGB}{155,0,255}

\newcommand{\argmax}{\operatornamewithlimits{argmax}}
\newcommand{\argmin}{\operatornamewithlimits{argmin}}
\newcommand{\nosection}[1]{\vspace{2mm}\noindent\textbf{#1}}
\newcommand{\minus}{\scalebox{0.5}[1.0]{$-$}}

\ifcvprfinal\fi
\begin{document}

\title{Near-Online Multi-target Tracking with Aggregated Local Flow Descriptor}

\author{Wongun Choi\\
NEC Laboratories America\\
10080 N. Wolfe Rd, Cupertino, CA, USA\\
{\tt\small wongun@nec-labs.com}
}

\maketitle

\begin{abstract}
In this paper, we focus on the two key aspects of multiple target tracking problem: 1) designing an accurate affinity measure to associate detections and 2) implementing an efficient and accurate (near) online multiple target tracking algorithm. As the first contribution, we introduce a novel Aggregated Local Flow Descriptor (ALFD) that encodes the relative motion pattern between a pair of temporally distant detections using long term interest point trajectories (IPTs). Leveraging on the IPTs, the ALFD provides a robust affinity measure for estimating the likelihood of matching detections regardless of the application scenarios. As another contribution, we present a Near-Online Multi-target Tracking (NOMT) algorithm. The tracking problem is formulated as a data-association between targets and detections in a temporal window, that is performed repeatedly at every frame. While being efficient, NOMT achieves robustness via integrating multiple cues including ALFD metric, target dynamics, appearance similarity, and long term trajectory regularization into the model. Our ablative analysis verifies the superiority of the ALFD metric over the other conventional affinity metrics. We run a comprehensive experimental evaluation on two challenging tracking datasets, KITTI~\cite{Geiger2012CVPR} and MOT~\cite{MOTChallenge} datasets. The NOMT method combined with ALFD metric achieves the best accuracy in both datasets with significant margins (about $10\%$ higher MOTA) over the state-of-the-arts.
\end{abstract}

\section{Introduction}
 
The goal of multiple target tracking is to automatically identify objects of interest and reliably estimate the motion of targets over the time. 
Thanks to the recent advancement in 
image-based object detection methods~\cite{DollarPAMI14pyramids,felzenszwalb2010object,girshick2013rich,wang2013regionlets}, \emph{tracking-by-detection}~\cite{BerclazFTF11,BreitensteinICCV09,Ess_PAMI_09,Kuo_CVPR_10,Milan:2014:CEM} has become a popular framework to tackle the multiple target tracking problem. The advantages of the framework are that it naturally identifies new objects of interest entering the scene, that it can handle video sequences recorded using mobile platforms, and that it is robust to a target drift. The key challenge in this framework is to accurately group the detections into individual targets with high accuracy (\emph{data association}), so one target could be fully represented by a single estimated trajectory. Mistakes made in the identity maintenance could result in a catastrophic failure in many high level reasoning tasks, such as future motion prediction, target behavior analysis, etc. 
\begin{figure}
\begin{center}
\includegraphics[width=\linewidth,trim=15mm 125mm 140mm 10mm,clip]{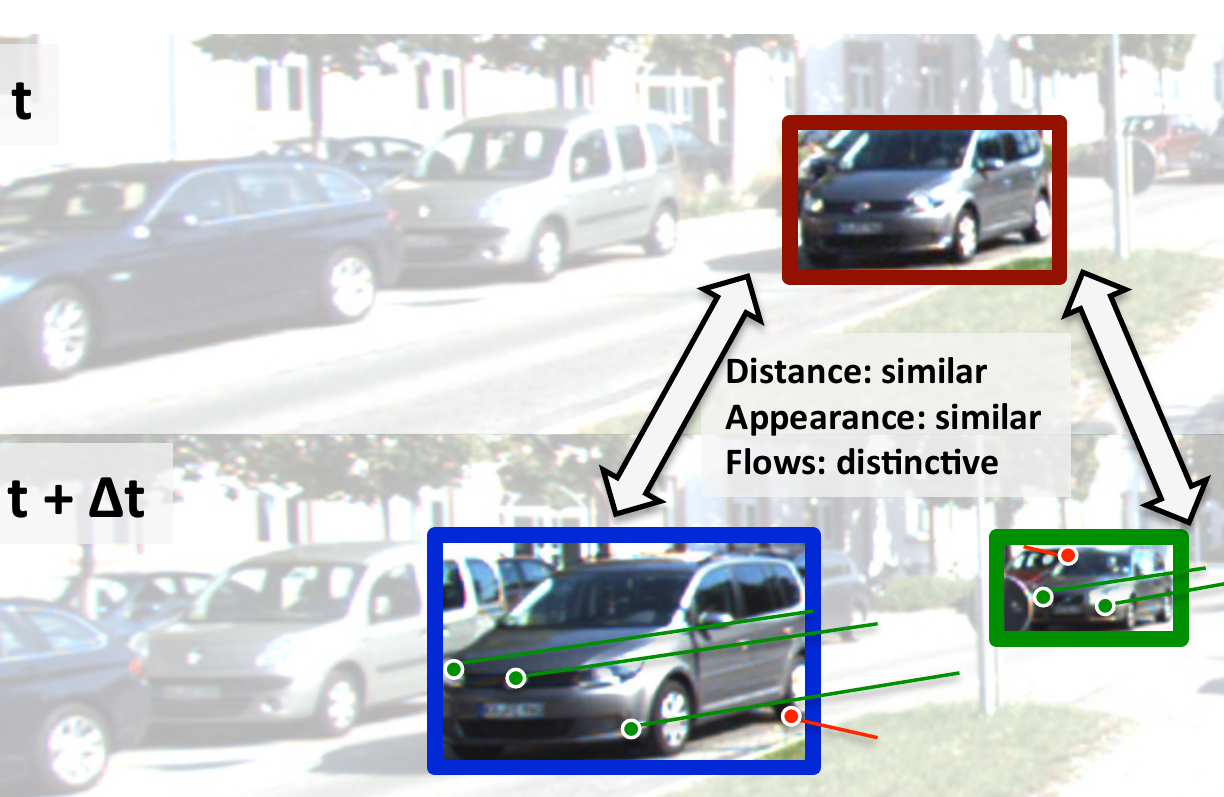}
\end{center}
\caption{Bounding box distance and appearance similarity are popularly used affinity metrics in the multiple target tracking literature. However, in real-world crowded scenes, they are often ambiguous to successfully distinguish adjacent or similar looking targets. Yet, the optical flow trajectories provide more reliable measure to compare different detections across time. Although individual trajectory may be inaccurate (\textcolor{red}{red} line), collectively they provide strong information to measure the affinity. We propose a novel Aggregated Local Flow Descriptor that exploits the optical flow reliably in the multiple target tracking problem. The figure is best shown in color.}
\label{fig:intro}
\end{figure}

\begin{figure*}
\begin{center}
\begin{tabular}{|@{\hspace{0.5mm}}c@{\hspace{0.5mm}}|@{\hspace{0.5mm}}c@{\hspace{0.5mm}}|@{\hspace{0.5mm}}c@{\hspace{0.5mm}}|}
\hline
at $t_1$ & at $t_2$ & at $t_3$\\
\hline
\includegraphics[width=0.32\linewidth,trim=18mm 18mm 40mm 50mm,clip]{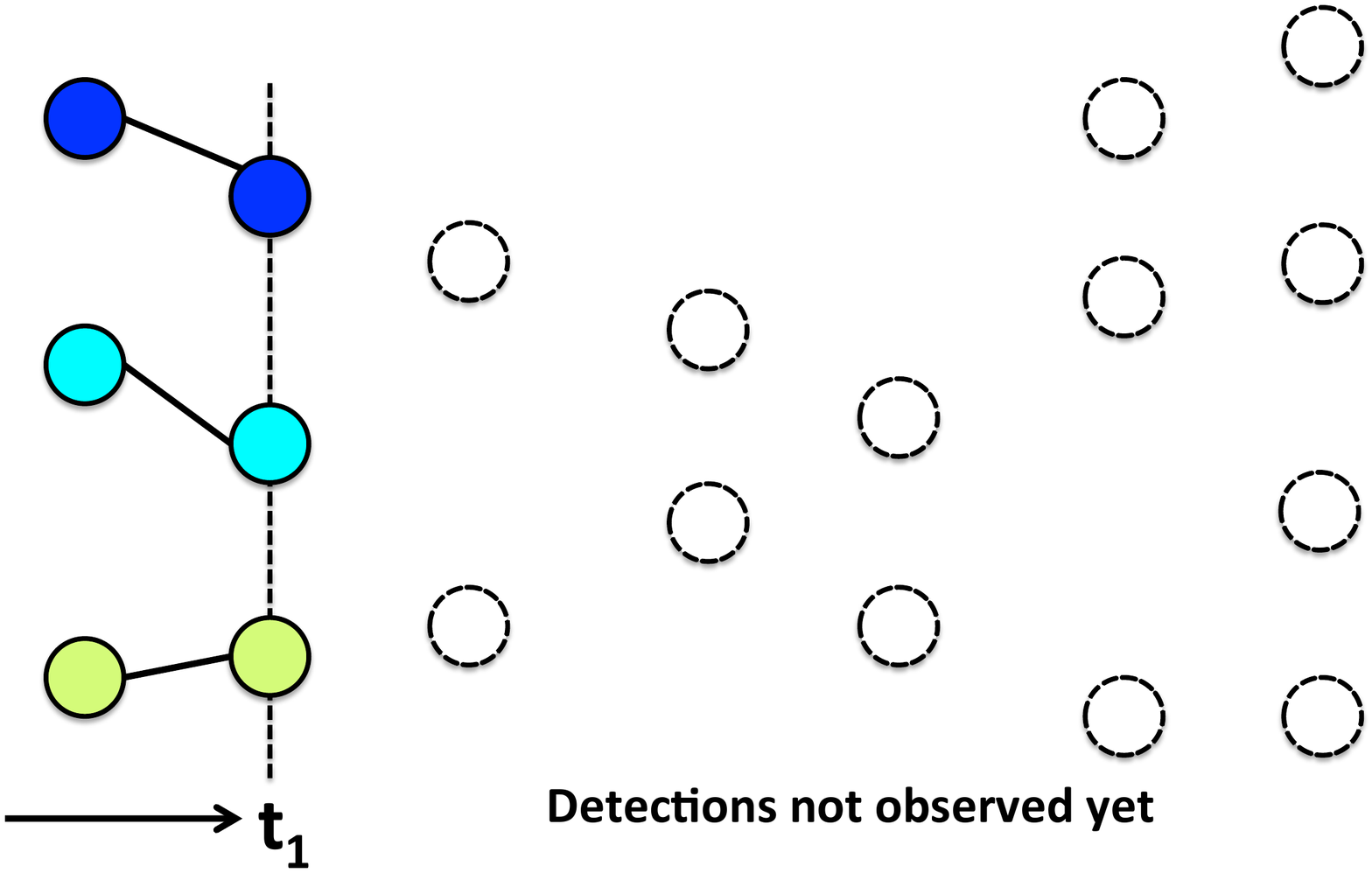} &
\includegraphics[width=0.32\linewidth,trim=18mm 18mm 40mm 50mm,clip]{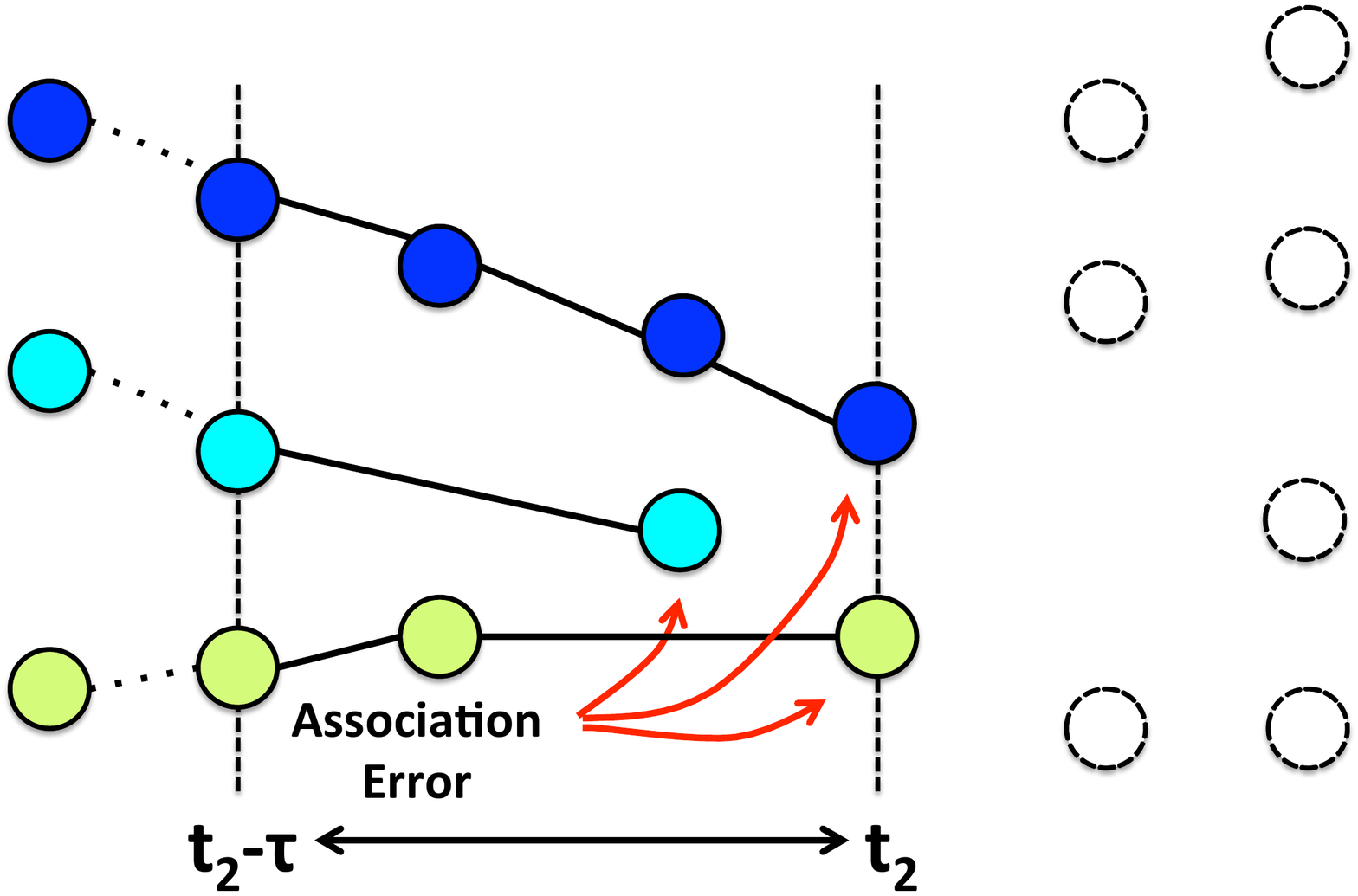} & 
\includegraphics[width=0.32\linewidth,trim=18mm 18mm 40mm 50mm,clip]{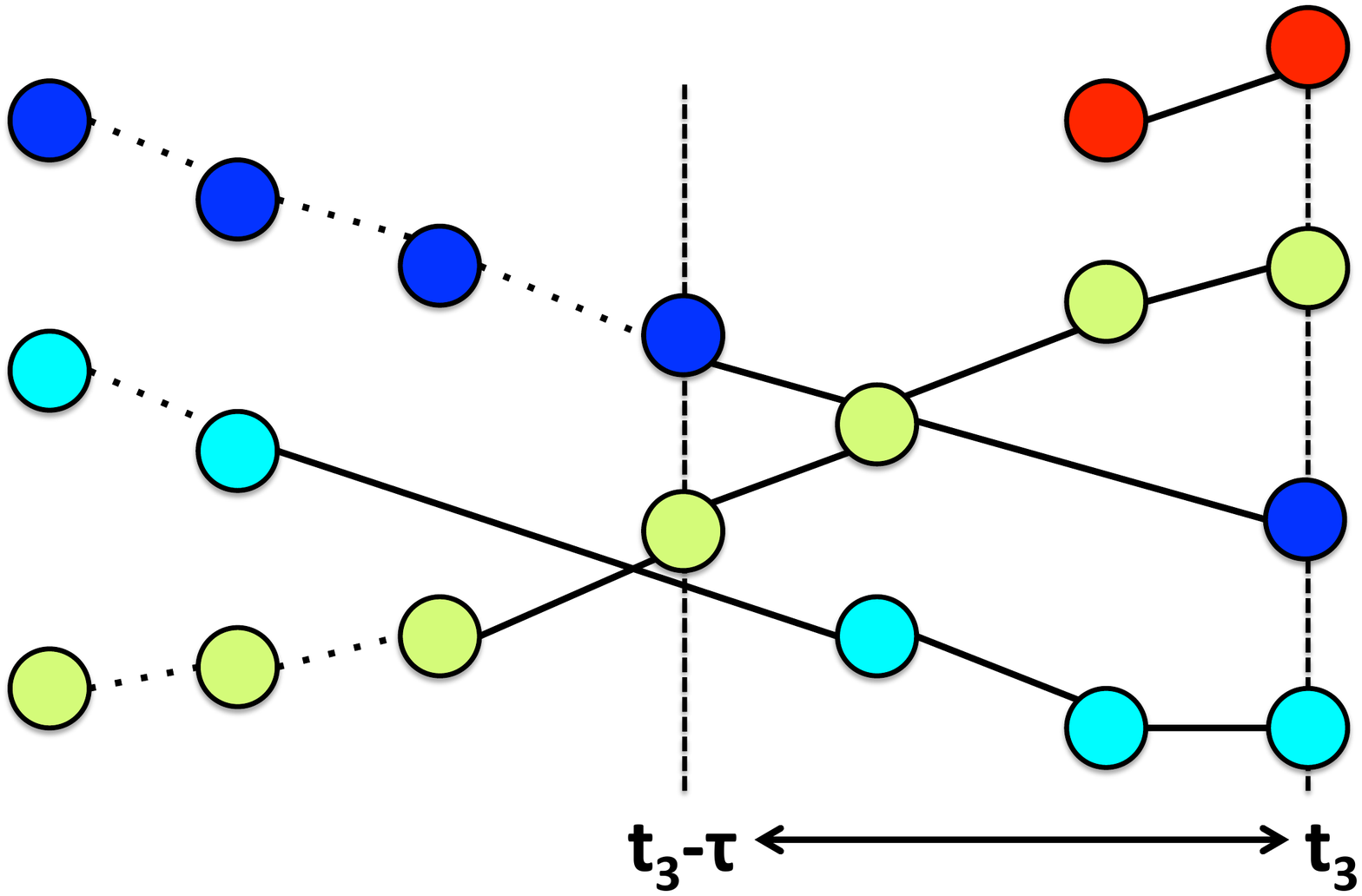}\\
\hline
\end{tabular}
\end{center}
\caption{Our NOMT algorithm solves the global association problem at every time frame $t$ with a temporal window $\tau$. Solid circles represent associated targets, dashed circles represent unobserved detections, dashed lines show finalized target association before the temporal window, and solid lines represent the (active) association made in the current time frame. Due to the limited amount of observation, the tracking algorithm may produce an erroneous association at $t_2$. But once more observation is provided at $t_3$, our algorithm is capable of fixing the error made in $t_2$. In addition, our method automatically identifies new targets on the fly (\textcolor{red}{red} circles). The figure is best shown in color.}
\label{fig:intro2}
\end{figure*}

To implement a highly accurate multiple target tracking algorithm, it is important to have a robust data association model and an accurate measure to compare two detections across time (pairwise affinity measure). Recently, much work is done in the design of the data association algorithm using global (batch) tracking framework~\cite{BerclazFTF11,Kuo_CVPR_10,Milan:2014:CEM,Zhang_CVPR_08}. 
Compared to the online counterparts~\cite{BreitensteinICCV09,choi_pami13,Ess_PAMI_09,Khan_PAMI_05}, these methods have a benefit of considering all the detections over entire time frames. With a help of clever optimization algorithms, they achieve higher data association accuracy than traditional online tracking frameworks. However, the application of these methods is fundamentally limited to post-analysis of video sequences. 
On the other hand, the pairwise affinity measure is relatively less investigated in the recent literature despite its importance. Most methods adopt weak affinity measures (see Fig.~\ref{fig:intro}) to compare two detections across time, such as spatial affinity (e.g. bounding box overlap or euclidean distance~\cite{Andriyenko:2012:DCO,BerclazFTF11,Pirsiavash_CVPR_11}) or simple appearance similarity (e.g. intersection kernel with color histogram~\cite{ZamirECCV12}). In this paper, we address the two key challenging questions of the multiple target tracking problem: 1) how to accurately measure the pairwise affinity between two detections (i.e. likelihood to link the two) and 2) how to efficiently apply the ideas in global tracking algorithms into an online application. 

As the first contribution, we present a novel \emph{Aggregated Local Flow Descriptor} (ALFD) that encodes the relative motion pattern between two detection boxes in different time frames (Sec.~\ref{sec:ALFD}). By aggregating multiple local interest point trajectories (IPTs), the descriptor encodes how the IPTs in a detection moves with respect to another detection box, and vice versa. The main intuition is that although each individual IPT may have an error, \emph{collectively} they provide a strong information for comparing two detections. With a learned model, we observe that ALFD provides strong affinity measure, thereby providing strong cues for the association algorithm. 
 
As the second contribution, we propose an efficient \emph{Near-Online Multi-target Tracking} (NOMT) algorithm. Incorporating the robust ALFD descriptor as well as long-term motion/appearance models motivated by the success of modern batch tracking methods, the algorithm produces highly accurate trajectories, while preserving the causality property and running in real-time ($\sim 10$ FPS). In every time frame $t$, the algorithm solves the global data association problem between targets and all the detections in a temporal window $[t\minus\tau, t]$ of size $\tau$ (see Fig.~\ref{fig:intro2}). The key property is that the algorithm is able to fix any association error made in the past when more detections are provided. In order to achieve both accuracy and efficiency, the algorithm generates candidate hypothetical trajectories using ALFD driven tracklets and solve the association problem with a parallelized junction tree algorithm (Sec.~\ref{sec:method}). 

We perform a comprehensive experimental evaluation on two challenging datasets: KITTI~\cite{Geiger2012CVPR} and MOT Challenge~\cite{MOTChallenge} datasets. The proposed algorithm achieves the best accuracy with a large margin over the state-of-the-arts (including batch algorithms) in both datasets, demonstrating the superiority of our algorithm. The rest of the paper is organized as follows. Sec.~\ref{sec:background} discusses the background and related work in multiple target tracking literature. Sec.~\ref{sec:ALFD} describes our newly proposed ALFD. Sec.~\ref{sec:method} presents overview of NOMT data association model and the algorithm. Sec.~\ref{sec:details} discusses the details of model design. We show the analysis and experimental evaluation in Sec.~\ref{sec:exp}, and finally conclude with Sec.~\ref{sec:conc}.

\section{Background}
\label{sec:background}

Given a video sequence $V_1^T = \{I_1, I_2, ..., I_T\}$ of length $T$ and a set of detection hypotheses $\mathbb{D}_1^T = \{d_1, d_2, ..., d_N\}$, where $d_i$ is parameterized by the frame number $t_i$, a bounding box $(d_i[x], d_i[y], d_i[w], d_i[h])$\footnote{$[x], [y], [w], [h]$ operators represent the x, y, width and height value, respectively.}, and the score $s_i$, 
the goal of multiple target tracking is to find a coherent set of targets (associations) $\mathbb{A} = \{ A_1, A_2, ..., A_M\}$, where each target $A_m$ are parameterized by a set of detection indices (e.g. $A_1 = \{d_1, d_{10}, d_{23}\}$)  during the time of presence; i.e. $(V_1^T, \mathbb{D}_1^T) \rightarrow \mathbb{A}$. 

\subsection{Data Association Models}

Most of multiple target tracking algorithms/systems can be classified into two categories: online method and global (batch) method. 

Online algorithms~\cite{BreitensteinICCV09,choi_pami13,Ess_PAMI_09,Khan_PAMI_05,pellegrini2009you} are formulated to find the association between existing targets and detections in the current time frame: $(V_t^t, \mathbb{D}_t^t, \mathbb{A}^{t-1}) \rightarrow \mathbb{A}^t$. The advantages of online formulation are: 1) it is applicable to online/real-time scenario and 2) it is possible to take advantage of targets' dynamics information available in $\mathbb{A}^{t-1}$. Such methods, however, are often prone to association errors since they consider only one frame when making the association. 
Solving the problem based on (temporally) local information can fundamentally limit the association accuracy. 
To avoid such errors, \cite{BreitensteinICCV09} adopts conservative association threshold together with detection confidence maps, or \cite{choi_pami13,Khan_PAMI_05,pellegrini2009you} model interactions between targets. 

Recently, global algorithms~\cite{Andriyenko:2012:DCO,BerclazFTF11,Milan:2014:CEM,Pirsiavash_CVPR_11,Zhang_CVPR_08} became much popular in the community, as more robust association is achieved when considering long-term information in the association process. One common approach is to formulate the tracking as the network flow problem to directly obtain the targets from detection hypothesis~\cite{BerclazFTF11,Pirsiavash_CVPR_11,Zhang_CVPR_08}; i.e. $(V_1^T, \mathbb{D}_1^T) \rightarrow \mathbb{A}^T$. Although they have shown promising accuracy in multiple target tracking, the methods are often over-simplified for the tractability concern. They ignore useful target level information, such as target dynamics and interaction between targets (occlusion, social interaction, etc). Instead of directly solving the problem at one step, other employ an iterative algorithm that progressively refines the target association~\cite{Andriyenko:2012:DCO,huang2008robust,Kuo_CVPR_10,Milan:2014:CEM}; i.e. $(V_1^T, \mathbb{D}_1^T, \mathbb{A}_{i}^T) \rightarrow \mathbb{A}_{i+1}^T$, where $i$ represent an iteration. Starting from short trajectories (tracklet), \cite{huang2008robust,Kuo_CVPR_10} associate them into longer targets in a hierarchical fashion. \cite{Andriyenko:2012:DCO,Milan:2014:CEM} iterate between two modes, association and continuous estimation. Since these methods obtain intermediate target information, targets' dynamics, interaction and high-order statistics on the trajectories could be accounted that can lead to a better association accuracy. However, it is unclear how to seamlessly extend such models to an online application.

We propose a novel framework that can fill in the gap between the online and global algorithms. The task is defined as to solve the following problem: $(V_{1}^t, \mathbb{D}_{t-\tau}^t, \mathbb{A}^{t-1}) \rightarrow \mathbb{A}^t$ in each time frame $t$, where $\tau$ is pre-defined temporal window size. Our algorithm behaves similar to the online algorithm in that it outputs the association in every time frame. 
The critical difference is that any decision made in the past is subject to change once more observations are available. The association problems in each temporal window are solved using a newly proposed global association algorithm. Our method is also reminiscent of iterative global algorithm, since we augment all the track iteratively (one iteration per frame) considering multiple frames, that leads to a better association accuracy. 

\subsection{Affinity Measures in Visual Tracking}
The importance of a robust pairwise affinity measure (i.e. likelihood of $d_i$ and $d_j$ being the same target) is relatively less investigated in the multi-target tracking literature. Most of the recent literature~\cite{Andriyenko:2012:DCO,BerclazFTF11,Pirsiavash_CVPR_11,ZamirECCV12} employs a spatial distance and/or an appearance similarity with simple features (such as color histograms). In order to learn a discriminative affinity metric, Kuo \emph{et al.}~\cite{Kuo_CVPR_10} introduces an online appearance learning with boosting algorithm using various feature inputs such as HoG~\cite{dalal2005histograms}, texture feature, and RGB color histogram. Milan \emph{et al.}~\cite{Milan:2014:CEM} and Zamir \emph{et al.}~\cite{ZamirECCV12} proposed to use a global appearance consistency measure to ensure a target has a similar (or smoothly varying) appearance over a long term. Although there have been many works exploiting appearance information or spatial smoothness, we are not aware of any work employing optical flow trajectories to define a likelihood of matching detections. Recently, Fragkiadaki \emph{et al.}~\cite{FragkiadakiZZS12} introduced a method to track multiple targets while jointly clustering optical flow trajectories. The work presents a promising result, but the model is complicated due to the joint inference on both target and flow level association. In contrast, our ALFD provides a strong pairwise affinity measure that is generally applicable in any tracking model.

\section{Aggregated Local Flow Descriptor}
\label{sec:ALFD}

\begin{figure}
\begin{center}
\includegraphics[width=\linewidth,trim=15mm 95mm 80mm 10mm,clip]{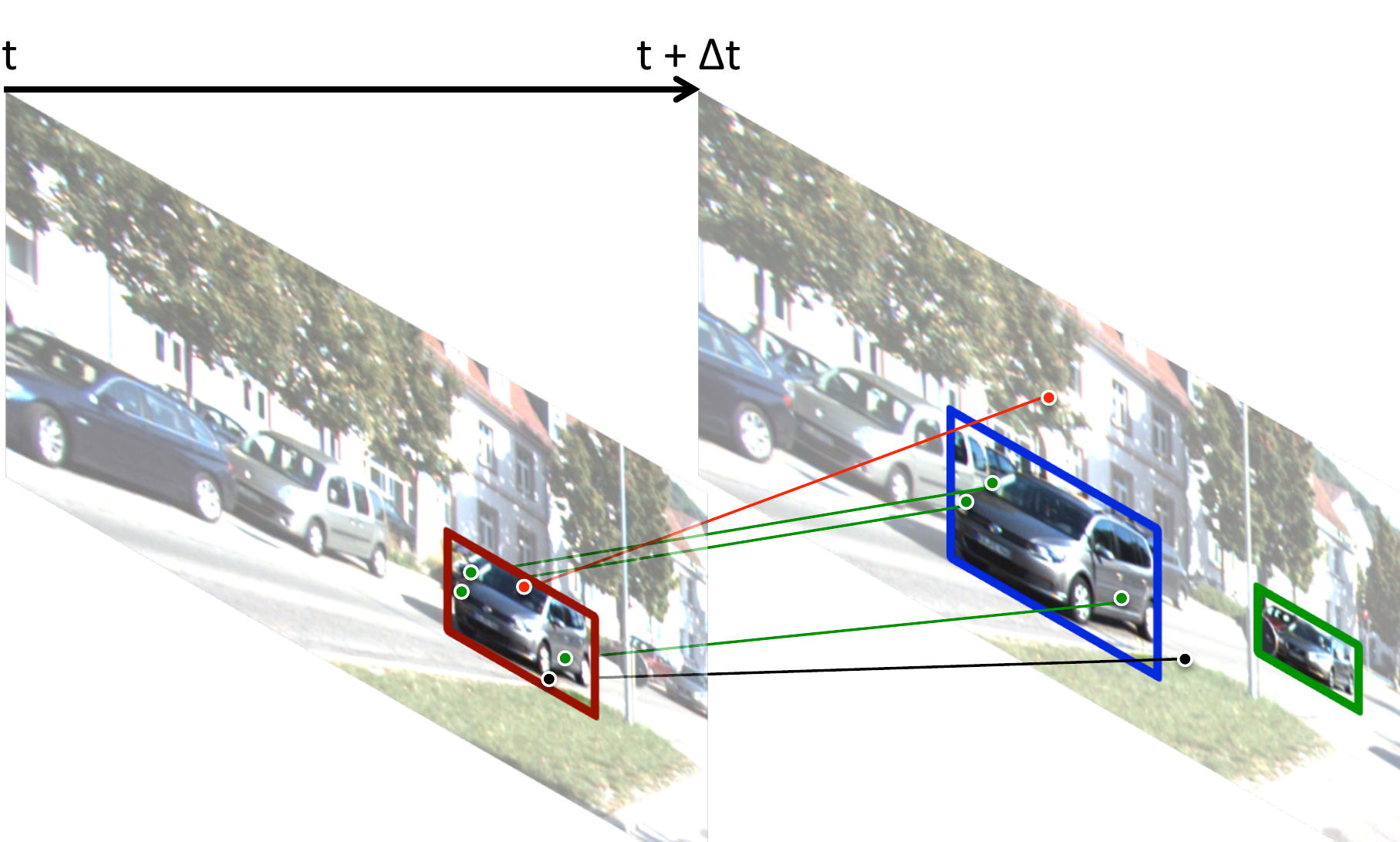}\\
\includegraphics[width=\linewidth,trim=10mm 76mm 30mm 10mm,clip]{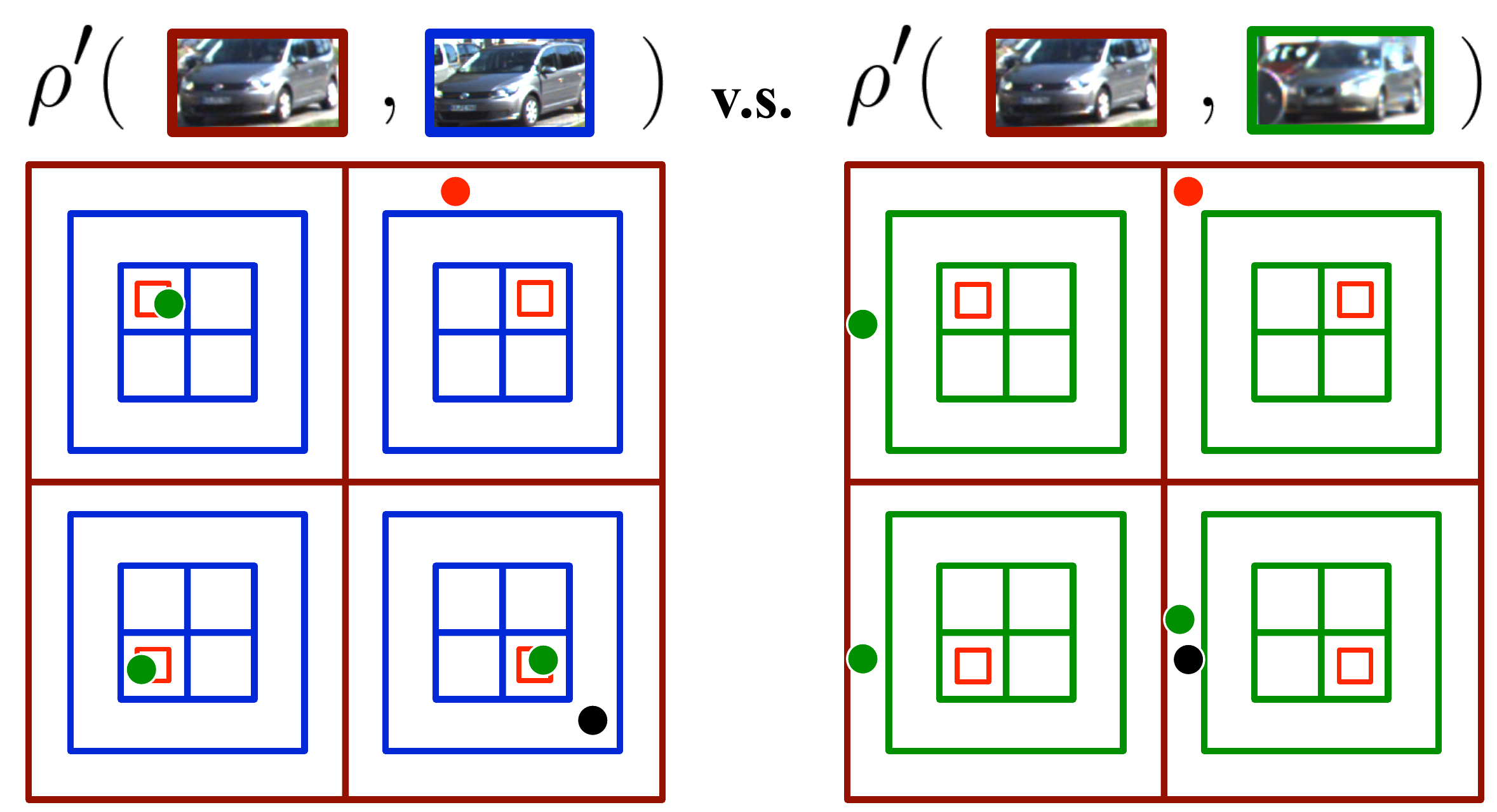}\\
\end{center}
\caption{Illustrative figure for unidirectional ALFDs $\rho'(d_i, d_j)$. In the top figure, we show detections as colored bounding boxes ($d_{\textcolor{red}{red}}$, $d_{\textcolor{blue}{blue}}$, and $d_{\textcolor{green}{green}}$). A pair of circles with connecting lines represent IPTs that are existing in both $t$ and $t + \triangle t$ and located inside of the $d_{\textcolor{red}{red}}$ at $t$. We draw the accurate (\textcolor{green}{green}), outlier (black), and erroneous (\textcolor{red}{red}) IPTs. In the bottom figure, we show two exemplar unidirectional ALFDs $\rho'$ for ($d_{\textcolor{red}{red}}$, $d_{\textcolor{blue}{blue}}$) and ($d_{\textcolor{red}{red}}$, $d_{\textcolor{green}{green}}$). The \textcolor{red}{red} grids ($2\times2$) represent the IPTs' location at $t$ relative to $d_{\textcolor{red}{red}}$. The \textcolor{blue}{blue} and \textcolor{green}{green} grids inside of each \textcolor{red}{red} bin ($2\times2 + 2$ external bins) shows the IPTs' location at $t+\triangle t$ relative to the corresponding boxes. IPTs in the grid bins with a \textcolor{red}{red} box are the one observed in the same relative location. Intuitively, the more IPTs are observed in the bins, the more likely the two detections belong to the same target. In contrast, wrong matches will have more supports in the outside bins. The illustration is shown using $2\times2$ grids to avoid clutter. We use $4\times4$ in practice. The figure is best shown in color.}
\label{fig:ALFDV2}
\end{figure}

The Aggregated Local Flow Descriptor (ALFD) encodes the relative motion pattern between two bounding boxes in a temporal distance ($\Delta t=|t_i-t_j|$) given interest point trajectories~\cite{Tomasi_CMUTR_91}. The main intuition in ALFD is that if the two boxes belong to the same target, we shall observe many supporting IPTs in the same relative location with respect to the boxes. In order to make it robust against small localization errors in detections, targets' orientation change, and outliers/errors in the IPTs, we build the ALFD using spatial histograms. Once the ALFD is obtained, we measure the affinity between two detections using the linear product of a learned model parameter $w_{\Delta t}$ and ALFD, i.e. $a_{A}(d_i, d_j) = w_{\Delta t} \cdot \rho(d_i, d_j)$. In the following subsections, we discuss the details of the design.

\subsection{Interest Point Trajectories}

We obtain Interest Point Trajectories using a local interest point detector~\cite{bradski2008learning,rosten2006machine} and optical flow algorithm~\cite{bradski2008learning,farneback2001very}. The algorithm is designed to produce a set of long and accurate point trajectories, combining various well-known computer vision techniques. Given an image $I_t$, we run the FAST interest point detector~\cite{bradski2008learning,rosten2006machine} to identify ``good points'' to track. In order to avoid having redundant points, we compute the distance between the newly detected interest points and the existing IPTs and keep the new points sufficiently far from the existing IPTs ($ > 4$ px). The new points are assigned unique IDs. For all the IPTs in $t$, we compute the forward ($t \rightarrow t+1$) and backward ($t+1 \rightarrow t$) optical flow using \cite{bradski2008learning,farneback2001very}. The starting points of backward flows are given by the forward flows' end point. Any IPT having a  large disagreement between the two ($> 10$ px) is terminated. 

\subsection{ALFD Design}

Let us define the necessary notations to discuss ALFD. $\kappa_{id} \in \mathcal{K}$ represents an IPT with a unique $id$ that is parameterized by pixel locations $(\kappa_{id}(t)[x], \kappa_{id}(t)[y])$ during the time of presence. $\kappa_{id}(t)$ denotes the pixel location at the frame $t$. If $\kappa_{id}$ does not exist at $t$ (terminated or not initiated), $\o$ is returned.

We first define a unidirectional ALFD $\rho'(d_i, d_j)$, i.e. from $d_i$ to $d_j$, by aggregating the information from all the IPTs that are located inside of $d_i$ box and existing at $t_j$. Formally, we define the IPT set as $\mathcal{K}(d_i, d_j) = \{\kappa_{id} | \kappa_{id}(t_i) \in d_i\ \&\ \kappa_{id}(t_j) \neq \o \}$. For each $\kappa_{id} \in \mathcal{K}(d_i, d_j)$, we compute the relative location $r_i(\kappa_{id}) = (x, y)$ of each $\kappa_{id}$ at $t_i$ by $r_i(\kappa_{id})[x] = ( \kappa_{id}(t_i)[x] \minus d_i[x] ) / d_i[w]$ and $r_i(\kappa_{id})[y] = ( \kappa_{id}(t_i)[y] \minus d_i[y] ) / d_i[h]$. We compute $r_j(\kappa_{id})$ similarly. Notice that $r_i(\kappa_{id})$ are bounded between $[0, 1]$, but $r_j(\kappa_{id})$ are not bounded since $\kappa_{id}$ can be outside of $d_j$. Given the $r_i(\kappa_{id})$ and $r_j(\kappa_{id})$, we compute the corresponding spatial grid bin indices as shown in the Fig.~\ref{fig:ALFDV2} and accumulate the count to build the descriptor. We define $4 \times 4$ grids for $r_i(\kappa_{id})$ and $4 \times 4 + 2$ grids for $r_j(\kappa_{id})$ where the last $2$ bins are accounting for the outside region of the detection. The first outside bin defines the neighborhood of the detection ($ < width/4\ \& < height/4$), and the second outside bin represents any farther region. 

Using a pair of unidirectional ALFDs, we define the ALFD as $\rho(d_i, d_j) = (\rho'(d_i, d_j) + \rho'(d_j, d_i))\ /\ n(d_i, d_j)$, where $n(d_i,d_j)$ is a normalizer. The normalizer $n$ is defined as $n(d_i, d_j) = | \mathcal{K}(d_i, d_j) | + | \mathcal{K}(d_j, d_i) | + \lambda$, where $| \mathcal{K}(\cdot) |$ is the count of IPTs and $\lambda$ is a constant. $\lambda$ ensures that the L1 norm of the ALFD increases as we have more supporting $\mathcal{K}(d_i, d_j)$ and converges to $1$. We use $\lambda=20$ in practice.

\begin{figure}[t!]
\begin{center}
\begin{tabular}{@{\hspace{0.5mm}}c@{\hspace{0.5mm}}c@{\hspace{0.5mm}}}
\includegraphics[width=0.48\linewidth,trim=50mm 88mm 35mm 82mm,clip]{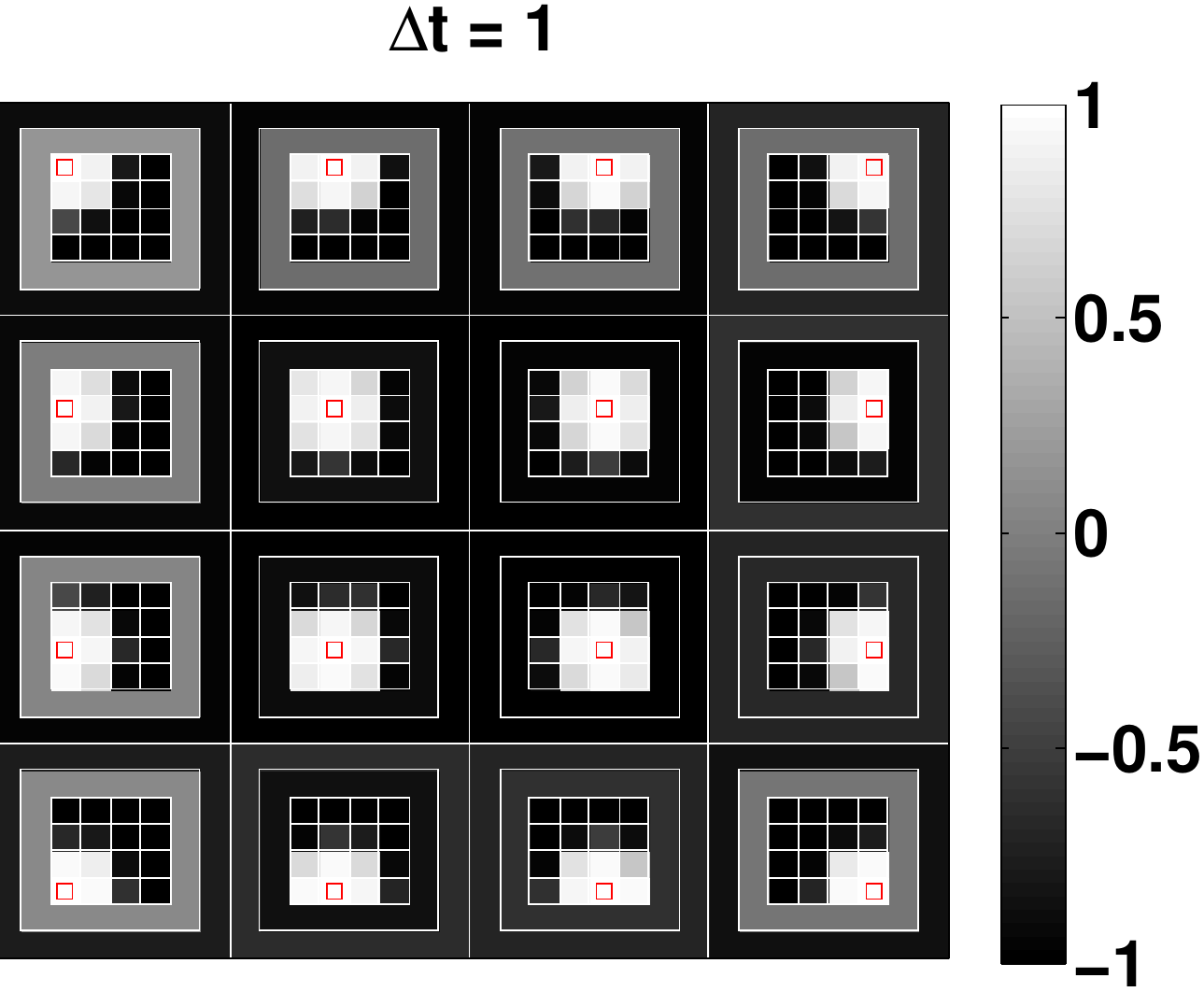} &
\includegraphics[width=0.48\linewidth,trim=50mm 88mm 35mm 82mm,clip]{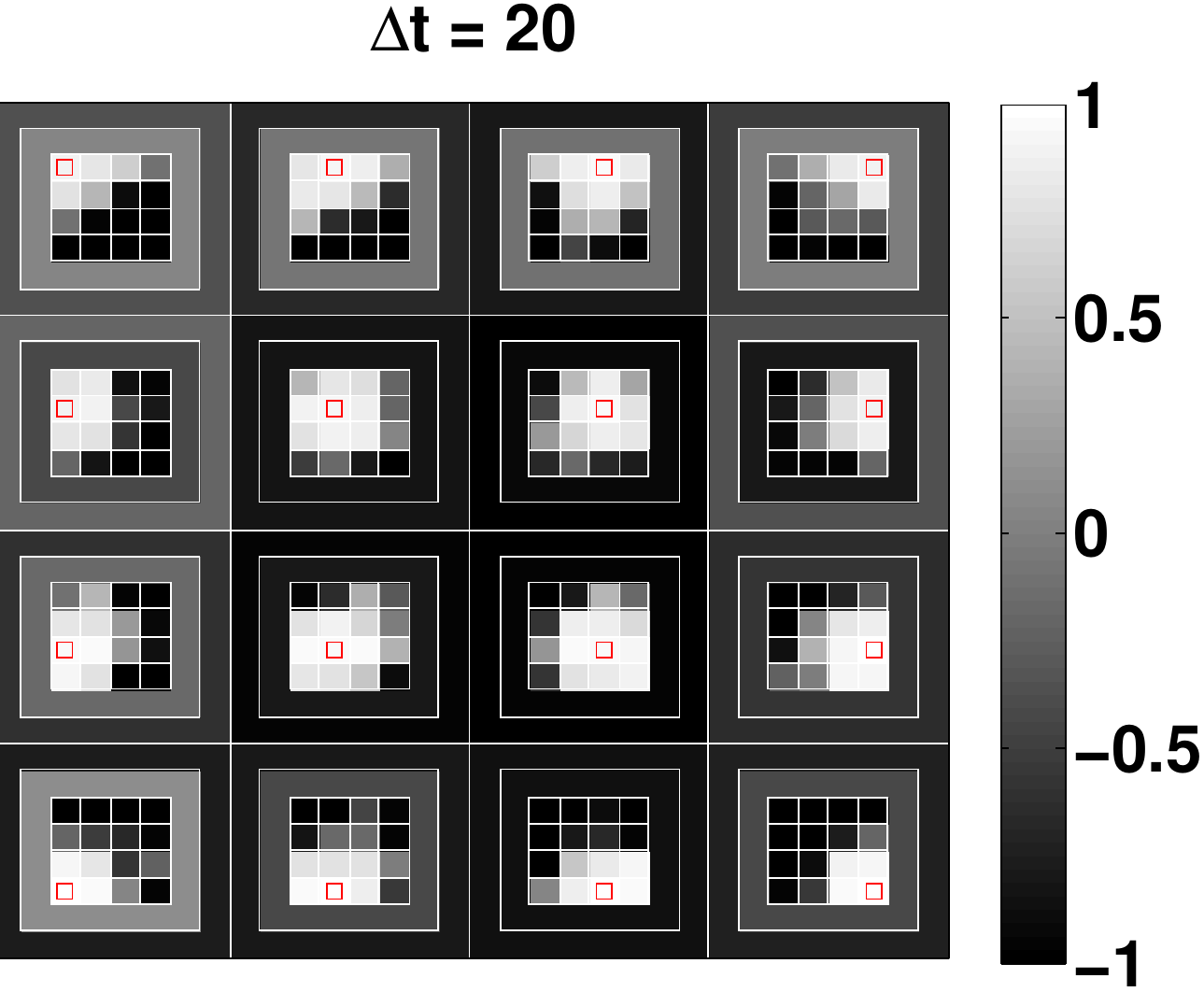} \\
\end{tabular}
\end{center}
\caption{Visualization of two learned model weights $w_{\Delta 1}$ and $w_{\Delta 20}$. Having a higher $\rho$ value in the bright (white) bins yields a higher affinity measure. As the temporal distance increase, the model weights tend to \emph{spread} out to the adjacent bins to account for a possible targets' orientation change and higher IPT errors. }
\label{fig:nomtmodels}
\end{figure}

\subsection{Learning the Model Weights}
We learn the model parameters $w_{\Delta t}$ from a training dataset with a weighted voting. Given a set of detections $\mathbb{D}_1^T$ and corresponding ground truth (GT) target annotations, we first assign the GT target id to each detections. For each detection $d_i$, we measure the overlap with all the GT boxes in $t_i$. If the best overlap $o_i$ is larger than $0.5$, the corresponding target id ($id_i$) is assigned. Otherwise, $-1$ is assigned. For all detections that has $id_i \geq 0$ (positive detections), we collect a set of detections $\mathcal{P}_i^{\Delta t} = \{d_j \in \mathbb{D}_1^T | t_j - t_i = \Delta t\}$. For each pair, we compute the margin $m_{ij}$ as follows: if $id_i$ and $id_j$ are identical, $m_{ij} = (o_i - 0.5) \cdot (o_j - 0.5)$. Otherwise, $m_{ij} = - (o_i - 0.5) \cdot (o_j - 0.5)$. Intuitively, $m_{ij}$ shall have a positive value if the two detections are from the same target, while $m_{ij}$ will have a negative value, if the $d_i$ and $d_j$ are from different targets. The magnitude is weighted by the localization accuracy. Given all the pairs and margins, we learn the model $w_{\Delta t}$ as follows:

{\scriptsize
\begin{equation}
w_{\Delta t} = \frac{\sum_{\{i \in \mathbb{D}_1^T | id_i \geq 0\}} \sum_{j \in \mathcal{P}_i^{\Delta t}} m_{ij} (\rho'(d_i, d_j) + \rho'(d_j, d_i))}{\sum_{\{i \in \mathbb{D}_1^T | id_i \geq 0\}} \sum_{j \in \mathcal{P}_i^{\Delta t}} |m_{ij}| (\rho'(d_i, d_j) + \rho'(d_j, d_i))}
\end{equation}}
The algorithm computes a weighted average with a sign over all the ALFD patterns, where the weights are determined by the overlap between targets and detections. Intuitively, the ALFD pattern between detections that matches well with GT contributes more on the model parameters. The advantage of the weighted voting method is that each element in $w_{\Delta t}$ are bounded in $[-1, 1]$, thus the ALFD metric, $a_{A}(d_i, d_j)$, is also bounded by $[-1, 1]$ since $||\rho(d_i, d_j)||_1 \leq 1$. Fig.~\ref{fig:nomtmodels} shows two learned model using our method. One can adopt alternative learning algorithms like SVM~\cite{CC01a}. 

\begin{figure*}[t!]
\begin{center}
\begin{tabular}{|@{\hspace{0.5mm}}c@{\hspace{0.5mm}}|@{\hspace{0.5mm}}c@{\hspace{0.5mm}}|@{\hspace{0.5mm}}c@{\hspace{0.5mm}}|@{\hspace{0.5mm}}c@{\hspace{0.5mm}}|}
\hline
(a) Inputs at $t$ & (b) Hypotheses Generation & (c) CRF Inference & (d) Outputs at $t$\\
\hline
\includegraphics[width=0.21\linewidth,trim=18mm 18mm 40mm 50mm,clip]{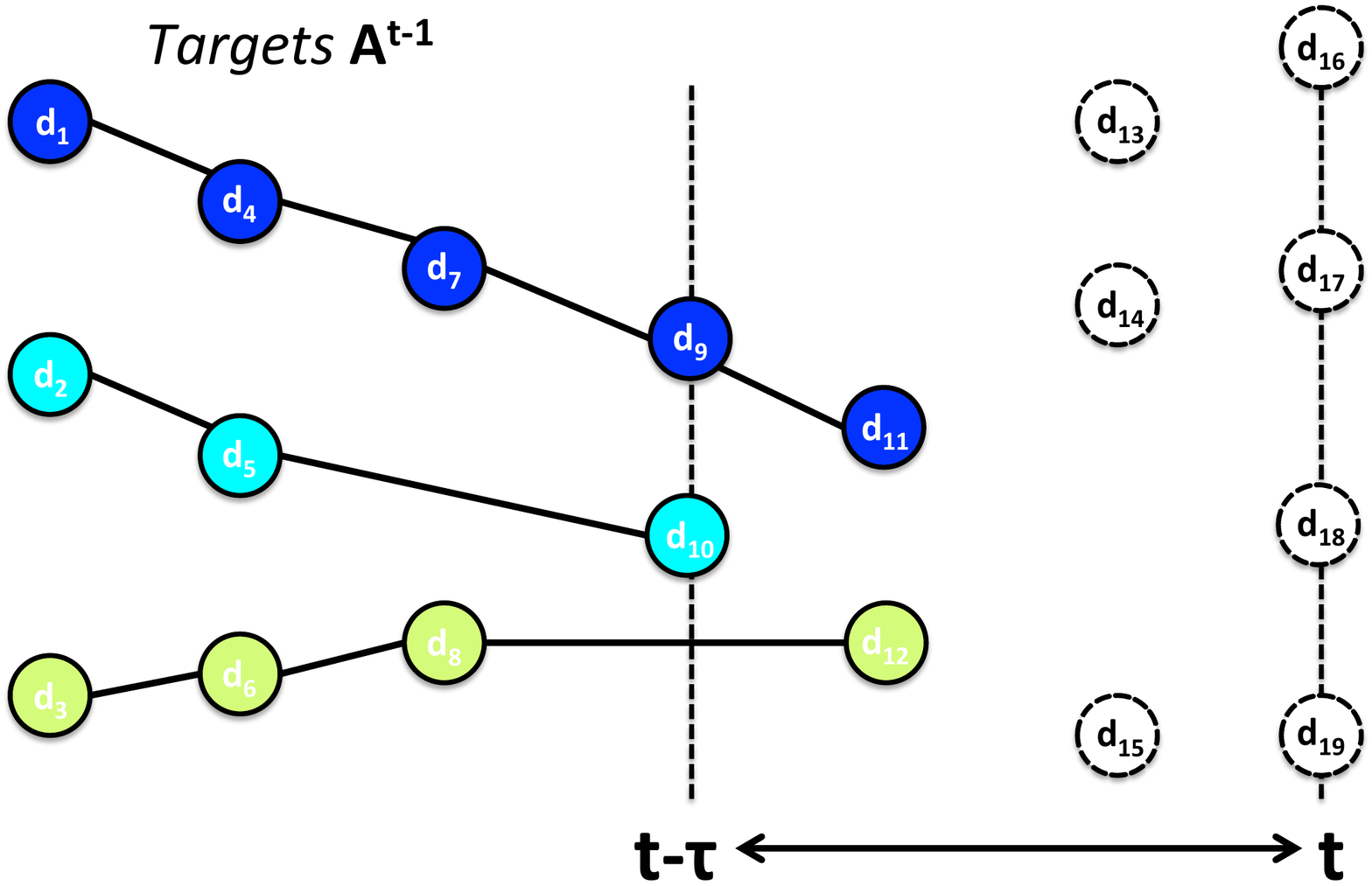} &
\includegraphics[width=0.28\linewidth,trim=13mm 60mm 15mm 30mm,clip]{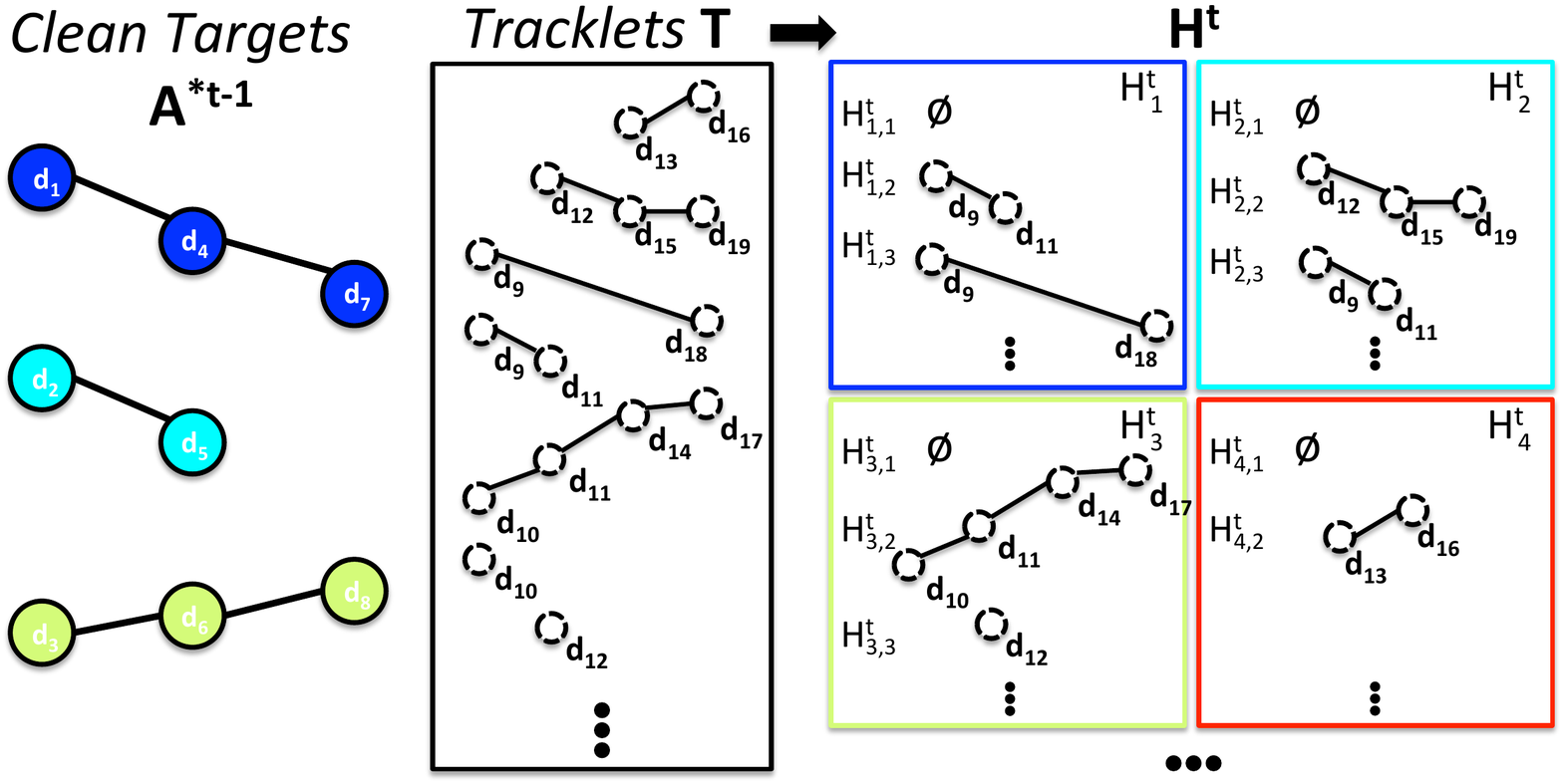} & 
\includegraphics[width=0.27\linewidth,trim=0mm 96mm 77mm 11mm,clip]{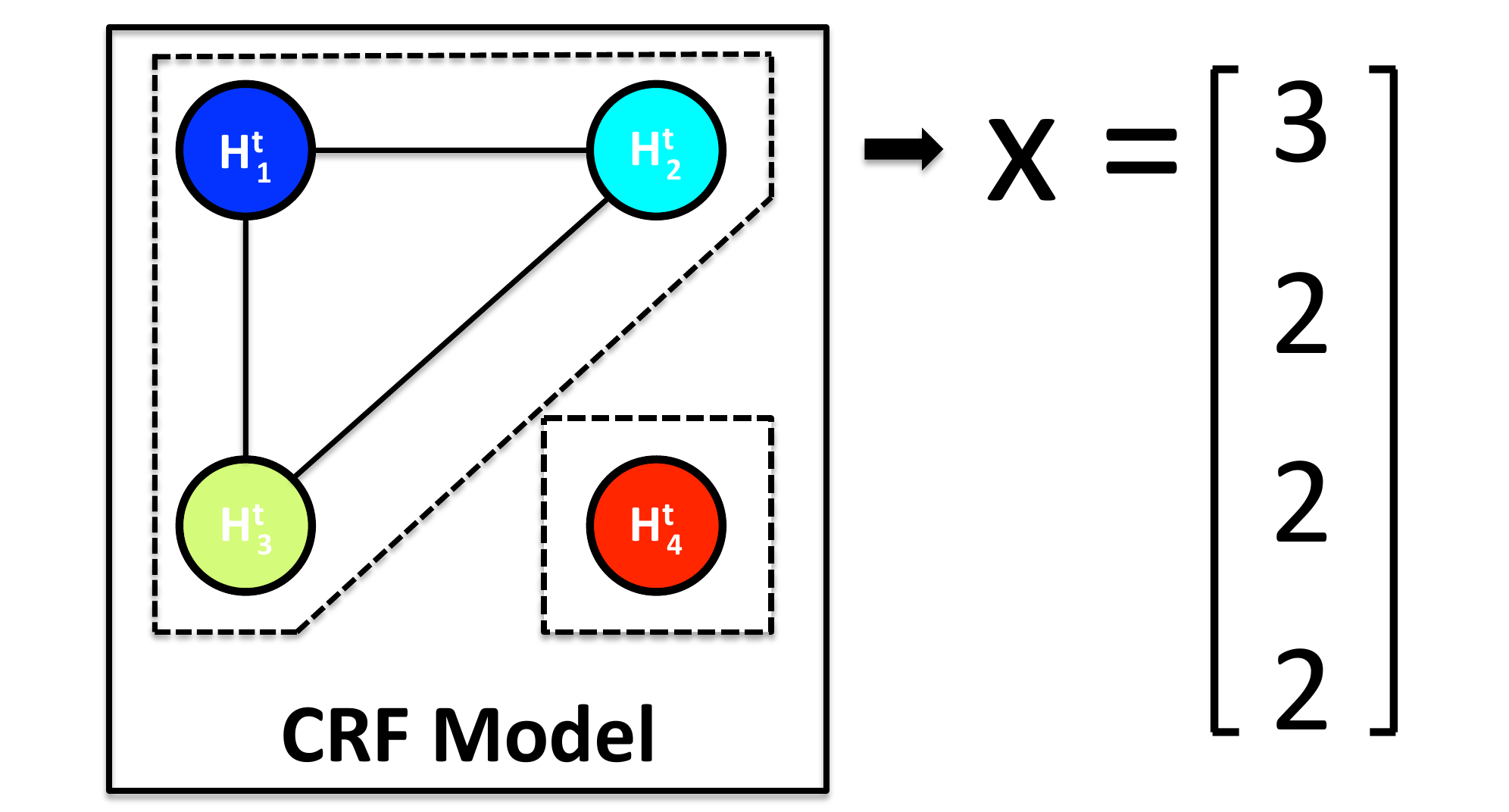} &
\includegraphics[width=0.21\linewidth,trim=18mm 18mm 40mm 50mm,clip]{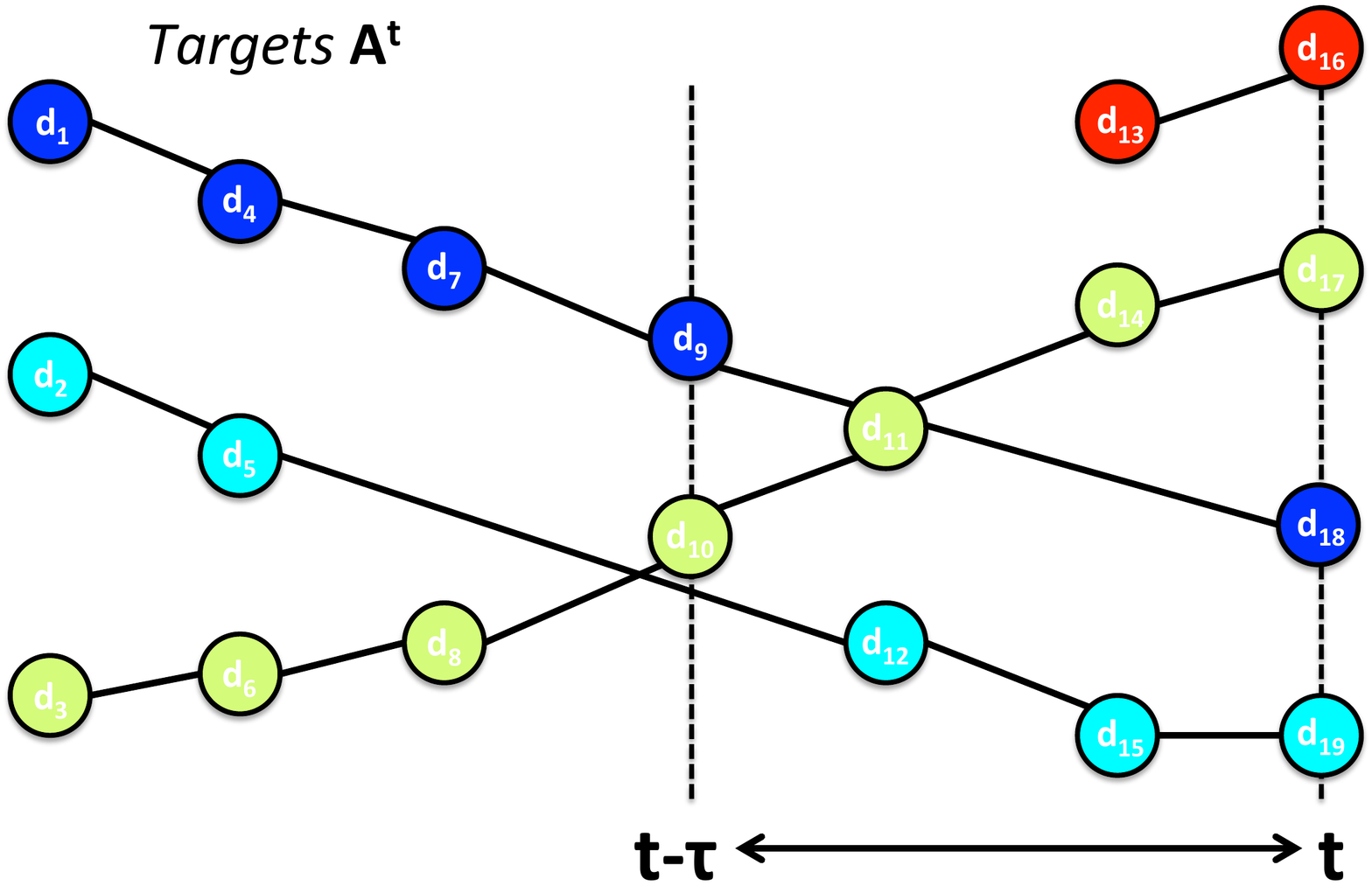}\\
\hline
\end{tabular}
\end{center}
\caption{
Schematic illustration of NOMT algorithm. (a) Given a set of existing targets $\mathbb{A}^{t-1}$ and detections $\mathbb{D}_{t - \tau}^t$, (b) our method generates a set of candidate hypotheses $\mathbb{H}^t$ using tracklets $\mathcal{T}$. Constructing a CRF model with the hypotheses, (c) we select the most consistent solution $x$ using our inference algorithm and (d) output targets $\mathbb{A}^{t}$ are obtained by augmenting previous targets $\mathbb{A}^{t-1}$ with the solution $\mathbb{H}^t(\hat{x})$. See text for the details.
}
\label{fig:nomtfig}
\end{figure*}

\subsection{Properties}
In this section, we discuss the properties of ALFD affinity metric $a_{A}(d_i, d_j)$. Firstly, unlike appearance or spatial metrics, ALFD implicitly exploit the information in all the images between $t_i$ and $t_j$ through IPTs. 
Secondly, thanks to the collective nature of ALFD design, it provides strong affinity metric over arbitrary length of time. We observe a significant benefit over the appearance or spatial metric especially over a long temporal distance (see Sec.~\ref{sec:exp:alfd} for the analysis). Thirdly, it is generally applicable to any scenarios (either static or moving camera) and for any object types (person or car). A disadvantage of the ALFD is that it may become unreliable when there is an occlusion. When an occlusion happens to a target, the IPTs initiated from the target tend to adhere to the occluder. It motivates us to combine target dynamics information discussed in Sec.~\ref{sec:unary}.


\section{Near Online Multi-target Tracking (NOMT)}
\label{sec:method}
We employ a near-online multi-target tracking framework that updates and outputs targets $\mathbb{A}^t$ in each time frame considering inputs in a temporal window $[t \minus \tau, t]$. We implement the NOMT algorithm with a hypothesis generation and selection scheme. 
For the convenience of discussion, we define \emph{clean} targets $\mathbb{A}^{* t \minus 1} = \{A_1^{* t \minus 1}, A_2^{* t \minus 1}, ...\}$ that exclude all the associated detections in $[t \minus \tau, t \minus 1]$.
Given a set of detections in $[t \minus \tau, t]$ and clean targets $\mathbb{A}^{* t \minus 1}$, we generate multiple target hypotheses $H_m^t = \{ H_{m, 1}^t = \o, H_{m, 2}^t, H_{m, 3}^t ... \}$ for each target $A_m^{*t-1}$ as well as newly entering targets, where $\o$ (empty hypothesis) represents the termination of the target and each $H_{m, k}^t$ indicates a set of candidate detections in $[t \minus \tau, t]$ that can be associated to a target  (Sec.~\ref{sec:generation}). Each $H_{m, k}^t$ may contain $0$ to $\tau$ detections (at one time frame, there can be $0$ or $1$ detection).
Given the set of hypotheses for all the existing and new targets, the algorithm finds the most consistent set of hypotheses (MAP) for all the targets (one for each) using a graphical model (sec.~\ref{sec:selection}). As the key characteristic, our algorithm is able to fix any association error (for the detections within the temporal window $[t \minus \tau, t]$ ) made in the previous time frames.

\subsection{Model Representation}
\label{sec:model}

Before going into the details of each step, we discuss our underlying model representation. The model is formulated as an energy minimization framework; $\hat{x} = \argmin_{x} E(\mathbb{A}^{* t \minus 1}, \mathbb{H}^t(x), \mathbb{D}_{t \minus \tau}^t, V_1^t)$, where $x$ is an integer state vector indicating which hypothesis is chosen for a corresponding target, 
$\mathbb{H}^t$ is the set of all the hypotheses $\{ H_1^t, H_2^t, ... \}$, and $\mathbb{H}^t(x)$ is a set of selected hypothesis $\{H_{1, x_1}^t, H_{2, x_2}^t, ... \}$. Solving the optimization, the updated targets $\mathbb{A}^t$ can be uniquely identified by augmenting $\mathbb{A}^{* t \minus 1}$ with the selected hypothesis $\mathbb{H}^t(\hat{x})$. Hereafter, we drop $V_1^t$ and $\mathbb{D}_{t \minus \tau}^t$ to avoid clutters in the equations. The energy is defined as follows:

{\footnotesize
\begin{align} 
E(\mathbb{A}^{* t \minus 1}, \mathbb{H}^t(x)) &= \sum_{m \in \mathbb{A}^{* t \minus 1}} \Psi(A_m^{* t \minus 1}, H_{m, x_m}^t) \nonumber \\
& + \sum_{ (m, l) \in \mathbb{A}^{* t \minus 1} } \Phi(H_{m, x_m}^t, H_{l, x_l}^t) \label{eq:obj}
\end{align}
}
where $\Psi( \cdot )$ encodes individual target's motion, appearance, and ALFD metric consistency, and $\Phi( \cdot )$ represent an exclusive relationship between different targets (e.g. no two targets share the same detection). If there are hypotheses for newly entering targets, we define the corresponding target as an empty set, $A_m^{* t-1} = \o$.

\nosection{Single Target Consistency}
\vspace{0.25em}

The potential measures the compatibility of a hypothesis $H_{m, x_m}^t$ to a target $A_m^{* t \minus 1}$. Mathematically, this can be decomposed into unary, pairwise and high order terms as follows:

{\scriptsize
\begin{align} 
\Psi(A_m^{* t \minus 1}, H_{m, x_m}^{t}) & = \sum_{i \in H_{m, x_m}^t} \psi_u(A_m^{* t \minus 1}, d_i) \nonumber \\
& + \sum_{(i, j) \in H_{m, x_m}^t} \psi_p(d_i, d_j) + \psi_{h}(A_m^{* t \minus 1}, H_{m, x_m}^t ) \label{eq:targetconsist}
\end{align}
}
$\psi_u$ encodes the compatibility of each detection $d_i$ in the target hypothesis $H_{m, x_m}^{t}$ using the ALFD affinity metric and Target Dynamics feature (Sec.~\ref{sec:unary}). 
$\psi_p$ measures the pairwise compatibility (self-consistency of the hypothesis) between detections within $H_{m, x_m}^{t}$ (Sec.~\ref{sec:pairwise}) using the ALFD metric. 
Finally, $\psi_{h}$ implements a long-term smoothness constraint and appearance consistency 
(Sec.~\ref{sec:targethorder}). 

\nosection{Mutual Exclusion}
\vspace{0.25em}

This potential penalizes choosing two targets with large overlap in the image plane (repulsive force) as well as duplicate assignments of a detection. 
Instead of using ``hard'' exclusion constraints as in the Hungarian Algorithm~\cite{Kuhn_NRLQ_55}, we use ``soft'' cost function for flexibility and computational simplicity. If the single target consistency is strong enough, soft penalization cost could be overcome. Also, this formulation makes it possible to reuse popular graph inference algorithms discussed in Sec.~\ref{sec:selection}. The potential can be written as follows:

{\footnotesize
\begin{align}
\Phi(H_{m, x_m}^t , H_{l, x_l}^t ) &= \sum_{f=t \minus \tau}^t  \alpha \cdot o^2(d(H_{m,x_m}^t, f), d(H_{l, x_l}^t, f)) \nonumber \\
& + \beta \cdot \mathbb{I}(d(H_{m,x_m}^t, f), d(H_{l, x_l}^t, f))
\end{align} }
where $d(H_{m,x_m}^t, f)$ gives the associated detection of $H_{m,x_m}^t$ at time $f$ (if none, $\o$ is returned), $o^2(d_i, d_j) = 2 * IoU(d_i, d_j)^2$, and $\mathbb{I}(d_i, d_j)$ is an indicator function. 
The former penalizes having too much overlap between hypotheses and the later penalizes duplicate assignments of detections. We use $\alpha = 0.5$ and $\beta = 100$ (large enough to avoid duplicate assignments).

\subsection{Hypothesis Generation}
\label{sec:generation}

Direct optimization over the aforementioned objective function (eq.~\ref{eq:obj}) is infeasible since the space of $\mathbb{H}^t$ is huge in practice. To cope with the challenge, we first propose a set of candidate hypotheses $H_m$ for each target independently (Fig.~\ref{fig:nomtfig}(b)) and find a coherent solution (MAP) using a CRF inference algorithm (sec.~\ref{sec:selection}). As all the subsequent steps depend on the generated hypotheses, it is critical to have a comprehensive set of target hypotheses. We generates the hypotheses of existing and new targets using \emph{tracklets}. Notice that following steps could be done in parallel since we generate the hypotheses set per target independently.

\nosection{Tracklet Generation}
\vspace{0.25em}

For all the confident detections ($\forall d_i \in \mathbb{D}_{t \minus \tau}^t,\ s.t.\ s_i > 0$), we build a tracklet using the ALFD metric $a_{A}$. Starting from one detection tracklet $\mathcal{T}_i = \{d_i\}$, we grow the tracklet by greedily adding the best matching detection $d_k$ such that $k = \argmax_{k \in \mathbb{D}_{t \minus \tau}^t \backslash \mathcal{T}_i} max_{j \in \mathcal{T}_i} a_{A}(d_j, d_k)$, where $\mathbb{D}_{t \minus \tau}^t \backslash \mathcal{T}_i$ is the set of detections in $[t\minus\tau, t]$ excluding the frames already included in $\mathcal{T}_i$. If the best ALFD metric is lower than $0.4$ or $\mathcal{T}_i$ is full (has $\tau$ number of detections), the iteration is terminated. In addition, we also extracts the residual detections from each $A_m^{t \minus 1}$ in $[t \minus \tau, t]$ to obtain additional tracklets (i.e. $\forall m, A_m^{t \minus 1} \backslash A_m^{*t \minus 1}$). Since there can be identical tracklets, we keep only unique tracklets in the output set $\mathbb{T}$.

\nosection{Hypotheses for Existing Targets}
\vspace{0.25em}

We generate a set of target hypotheses $H_m^t$ for each existing target $A_m^{* t\minus 1}$ using the tracklets $\mathbb{T}$. In order to avoid having unnecessarily large number of hypotheses, we employ a gating strategy. For each target $A_m^{* t\minus 1}$, we obtain a target predictor using the least square algorithm with polynomial function~\cite{leon1980linear}. We vary the order of the polynomial depending on the dataset ($1$ for MOT and $2$ for KITTI). If there is an overlap (IoU) larger than a certain threshold between the prediction and the detections in the tracklet $\mathcal{T}_i$ at any frame in $[t \minus \tau, t]$, we add $\mathcal{T}_i$ to the hypotheses set $H_m^t$. In practice, we use a conservative threshold $0.1$ to have a rich set of hypotheses. Too old targets (having no associated detection in $[t \minus \tau \minus T_{active}, t]$) are ignored to avoid unnecessary computational burden. We use $T_{active} = 1\ sec$. 

\nosection{New Target Hypotheses}
\vspace{0.25em}

Since new targets can enter the scene at any time and at any location, it is desirable to automatically identify new targets. Our algorithm can naturally identify the new targets by treating any tracklet in the set $\mathbb{T}$ as a potential new target. We use a non-maximum suppression on tracklets to avoid having duplicate new targets. For each tracklet $\mathcal{T}_i$, we simply add an empty target $A_m^{* t\minus 1} = \o$ to $\mathbb{A}^{* t\minus 1}$ with an associated hypotheses set $H_m^t = \{\o, \mathcal{T}_i\}$.

\subsection{Inference with Dynamic Graphical Model}
\label{sec:selection}

Once we have all the hypotheses for all the new and existing targets, the problem (eq.~\ref{eq:obj}) can be formulated as an inference problem with an undirected graphical model, where one node represents a target and the states are hypothesis indices as shown in Fig.~\ref{fig:nomtfig} (c). The main challenges in this problem are: 1) there may exist loops in the graphical model representation and 2) the structure of graph is different depending on the hypotheses at each circumstance. In order to obtain the exact solution efficiently, we first analyze the structure of the graph on the fly and apply appropriate inference algorithms based on the structure analysis.

Given the graphical model, we find independent subgraphs (shown as dashed boxes in Fig.~\ref{fig:nomtfig} (c)) using connected component analysis~\cite{hopcroft1973algorithm} and perform individual inference algorithm per each subgraph in parallel. If a subgraph is composed of more than one node, we use junction-tree algorithm~\cite{koller2009probabilistic,LIBDAI} to obtain the solution for corresponding subgraph. Otherwise, we choose the best hypothesis for the target.

Once the states $x$ are found, we can uniquely identify the new set of targets by augmenting $\mathbb{A}^{* t \minus 1}$ with $\mathbb{H}^t(x)$:  $\mathbb{A}^{* t \minus 1} + \mathbb{H}^t(x) \rightarrow \mathbb{A}^t$. This process allows us to adjust any associations of $\mathbb{A}^{t \minus 1}$ in $[t\minus \tau, t]$ (i.e. addition, deletion, replacement, or no modification).

\section{Model Details}
\label{sec:details}
In this section, we discuss the details of the potentials described in the Eq.~\ref{eq:targetconsist}.

\subsection{Unary potential}
\label{sec:unary}

As discussed in the previous sections, we utilize the ALFD metric as the main affinity metric to compare detections. The unary potential for each detection in the hypothesis is measured by:

{\footnotesize
\begin{equation}
	\mu_A (A_m^{* t\minus 1}, d_i) = - \sum_{\Delta t \in \mathcal{N}} a_A(d(A_m^{* t\minus 1}, t_i - \Delta t), d_i)
\end{equation}}
where $\mathcal{N}$ is a predefined set of neighbor frame distances and $d(A_m^{* t\minus 1}, t_i)$ gives the associated detection of $A_m^{* t\minus 1}$ at $t_i$. Although we can define an arbitrarily large set of $\mathcal{N}$, we choose $\mathcal{N} = \{1, 2, 5, 10, 20\}$ for computational efficiency while modeling long term affinity measures. 

Although ALFD metric provides very strong information in most of the cases, there are few failure cases including occlusions, erroneous IPTs, etc. To complement such cases, we design an additional Target Dynamics (TD) feature $\mu_T (A_m^{* t\minus 1}, d_i)$. Using the same polynomial least square predictor discussed in Sec.~\ref{sec:generation}, we define the feature as follows:

{\footnotesize
\begin{equation}
\mu_T(A_m^{* t \minus 1}, d_i) = 
\left\{\begin{matrix}
\infty,\ \ \ \ if\ o^2(p(A_m^{* t \minus 1}, t_i), d_i) < 0.5\\ 
-\eta^{t_i - f(A_m^{* t \minus 1})} o^2(p(A_m^{* t \minus 1}, t_i), d_i),\ \ \ \ otherwise
\end{matrix}\right.
\end{equation}}
where $\eta$ is a decay factor ($0.98$) that discounts long term prediction, $f(A_m^{* t \minus 1})$ denotes the last associated frame of $A_m^{* t \minus 1}$, $o^2$ represents $IoU^2$ discussed in the Sec.~\ref{sec:model}, and $p$ is the polynomial least square predictor described in Sec.~\ref{sec:generation}. 

Using the two measures, we define the unary potential $\psi_u(A_m^{* t \minus 1}, d_i)$ as:

{\footnotesize
\begin{equation}
\psi_u(A_m^{* t \minus 1}, d_i) = \min(\mu_A(A_m^{* t \minus 1}, d_i), \mu_T(A_m^{* t \minus 1}, d_i)) - s_i
\end{equation}
}
where $s_i$ represents the detection score of $d_i$. The $min$ operator enables us to utilize the ALFD metric in most cases, but \emph{activate} the TD metric only when it is very confident (more than $0.5$ overlap between the prediction and the detection). If $A_m^{* t \minus 1}$ is empty, the potential becomes $-s_i$.

\subsection{Pairwise potential}
\label{sec:pairwise}

The pairwise potential $\psi_p(\cdot)$ is solely defined by the ALFD metric. Similarly to the unary potential, we define the pairwise relationship between detections in $H_{m, x_m}^t$, 
{\footnotesize
\begin{equation}
\psi_p (d_i, d_j) = 
\left\{\begin{matrix}
-a_A(d_i, d_j),\ \ \ if\ |d_i-d_j| \in \mathcal{N}\\ 
0,\ \ \ \ \ \ otherwise
\end{matrix}\right.
\end{equation}}
It measures the self-consistency of a hypothesis $H_{m, x_m}^t$.

\subsection{High-order potential}
\label{sec:targethorder}
We incorporate a high-order potential to regularize the target association process with a physical feasibility and appearance similarity.
Firstly, inspired by \cite{Andriyenko:2012:DCO,ZamirECCV12}, we implement the physical feasibility by penalizing the hypotheses that present an abrupt motion. Secondly, we encodes long term appearance similarity between all the detections in $A_m^{* t\minus 1}$ and $H_{m, x_m}^t$ similarly to \cite{ZamirECCV12}. 
The intuition is encoded by the following potential:

{\scriptsize
\begin{align}
\psi_h  (A_m^{* t\minus 1}, H_{m, x_m}^t) &=  \gamma \cdot \sum_{i \in H_{m, x_m}^t} \xi(p(A_m^{* t\minus 1} \cup H_{m, x_m}^t, t_i), d_i) \nonumber \\
& + \epsilon \cdot \sum_{(i, j) \in A_m^{* t\minus 1} \cup H_{m, x_m}^t} \theta - K(d_i, d_j)
\end{align}}
where $\gamma, \epsilon, \theta$ are scalar parameters, $\xi(a, b)$ measures the sum of squared distances in $(x, y, height)$ of the two boxes, that is normalized by the mean height of $p$ in $[t\minus \tau, t]$, and $K(d_i, d_j)$ represents the intersection kernel for color histograms associated with the detections. 
We use a pyramid of LAB color histogram where the first layer is the full box and the second layer is $3 \times 3$ grids. Only the A and B channels are used for the histogram with $4$ bins per each channel (resulting in $4 \times 4 \times (1+9)$ bins). We use $(\gamma, \epsilon, \theta) = (20, 0.4, 0.8)$ in practice. 

\section{Experimental Evaluation}
\label{sec:exp}
In order to evaluate the proposed algorithm, we use the KITTI object tracking benchmark~\cite{Geiger2012CVPR} and MOT challenge dataset~\cite{MOTChallenge}.
KITTI tracking benchmark is composed of about $19,000$ frames ($\sim32$ minutes). The dataset is composed of $21$ training and $29$ testing video sequences that are recorded using cameras mounted on top of a moving vehicle. Each video sequence has a variable number of frames from $78$ to $1176$ frames having a variable number of target objects (\emph{Car, Pedestrian, and Cyclist}). The videos are recorded at $10$ FPS. The dataset is very challenging since 1) the scenes are crowded (occlusion and clutter), 2) the camera is not stationary, and 3) target objects appears in arbitrary location with variable sizes. Many conventional assumptions/techniques adopted in multiple target tracking with a surveillance camera is not applicable in this case (e.g. fixed entering/exiting location, background subtraction, etc).
MOT challenge is composed of $11,286$ frames ($\sim16.5$ minutes) with varying FPS. The dataset is composed of $11$ training and $11$ testing video sequences. Some of the videos are recorded using mobile platform and the others are from surveillance videos. All the sequences contain only Pedestrians. As it is composed of videos with various configuration, tracking algorithms that are particularly tuned for a specific scenario would not work well in general.   
For the evaluation, we adopt the widely used CLEAR MOT tracking metrics~\cite{keni2008evaluating}. For a fair comparison to the other methods, we use the reference object detections provided by the both datasets. 

\subsection{ALFD Analysis}
\label{sec:exp:alfd}

\begin{table}[t]
\begin{center}
{\scriptsize
\begin{tabular}{| c || c | c | c | c | c |}
\hline
\multicolumn{6}{|c|}{KITTI 0001: Cars, Mobile camera} \\
\hline
Metric & $\triangle t = 1$& $\triangle t = 2$& $\triangle t = 5$& $\triangle t = 10$& $\triangle t = 20$ \\ 
\hline
ALFD & \textbf{0.91} & \textbf{0.89} & \textbf{0.84} & \textbf{0.80} & \textbf{0.71} \\
NDist2 & 0.81 & 0.66 & 0.32 & 0.15 & 0.06 \\
HistIK & 0.81 & 0.76 & 0.62 & 0.51 & 0.38  \\
\hline
\hline
\multicolumn{6}{|c|}{PETS09-S2L1: Pedestrians, Static camera} \\
\hline
Metric & $\triangle t = 1$& $\triangle t = 2$& $\triangle t = 5$& $\triangle t = 10$& $\triangle t = 20$ \\ 
\hline
ALFD & \textbf{0.88} & \textbf{0.86} & \textbf{0.83} & \textbf{0.78} & \textbf{0.68} \\
NDist2 & 0.85 & 0.78 & 0.67 & 0.55 & 0.41\\
HistIK & 0.76 & 0.71 & 0.65 & 0.60 & 0.51  \\
\hline
\end{tabular}
}
\caption{AUC of affinity metrics for varying $\triangle t$. Notice that ALFD provides a robust affinity metric even at $20$ frames distance. The results verify that ALFD provides stable affinity measure regardless of object type or the camera motion.}
\label{tab:alfd}
\end{center}

\end{table}

\begin{figure}[t]
\begin{center}
{\scriptsize
\begin{tabular}{c@{\hspace{1mm}}c@{\hspace{1mm}}c}
\multicolumn{3}{c}{KITTI 0001: Cars, Mobile camera} \\
\includegraphics[width=0.3\linewidth, trim =22mm 72mm 25mm 66mm, clip]{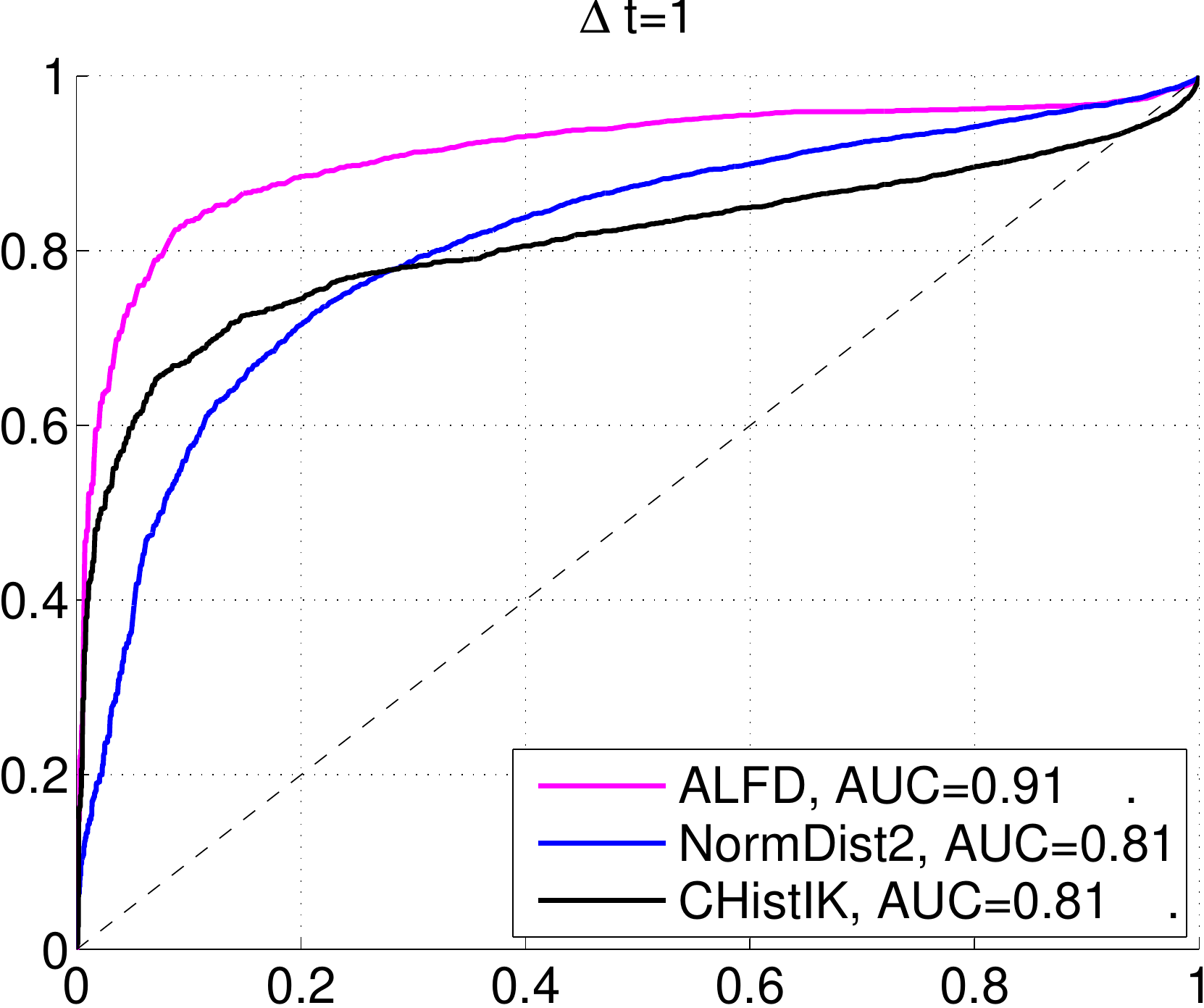} &
\includegraphics[width=0.3\linewidth, trim =22mm 72mm 25mm 66mm, clip]{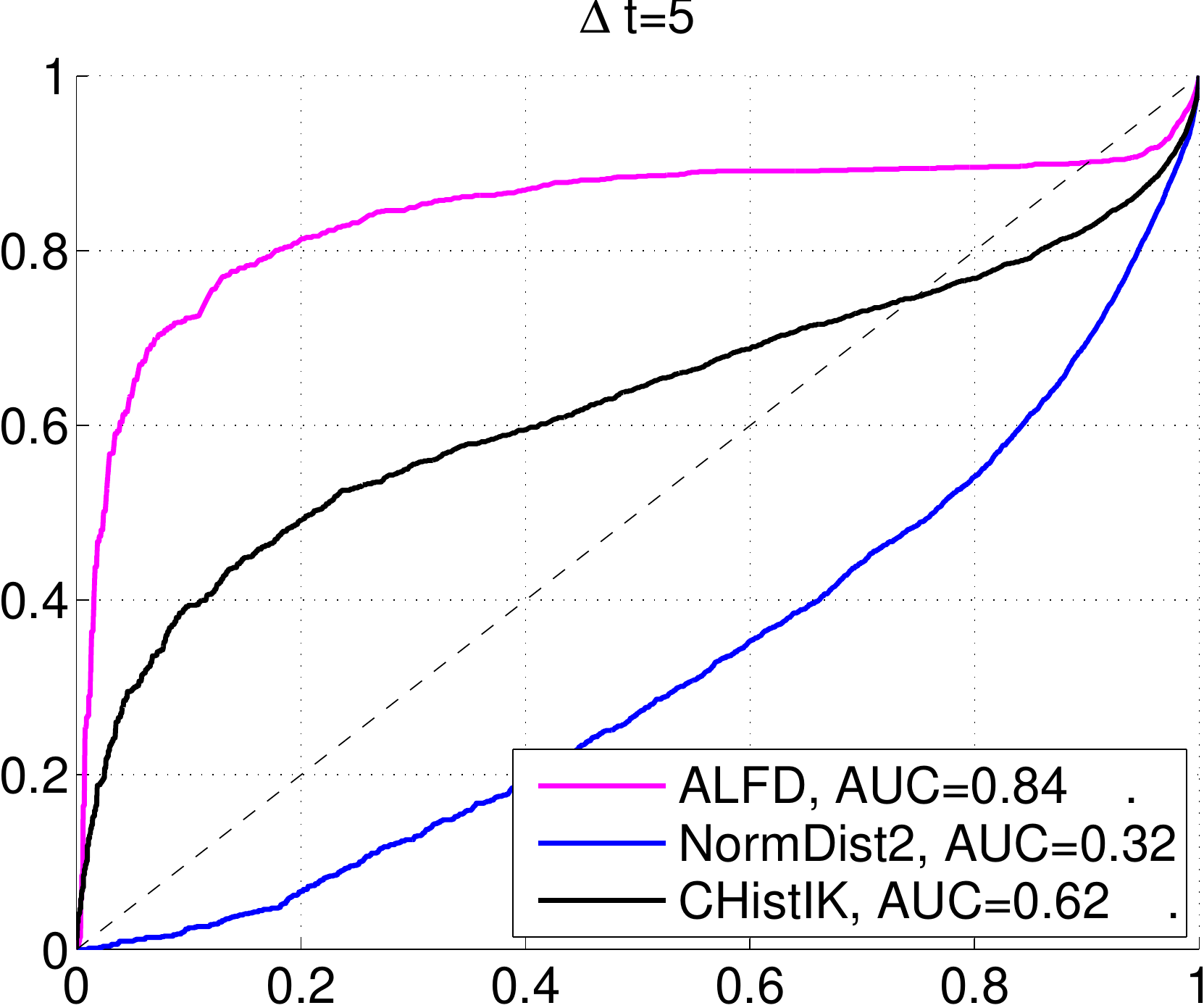} &
\includegraphics[width=0.3\linewidth, trim =22mm 72mm 25mm 66mm,  clip]{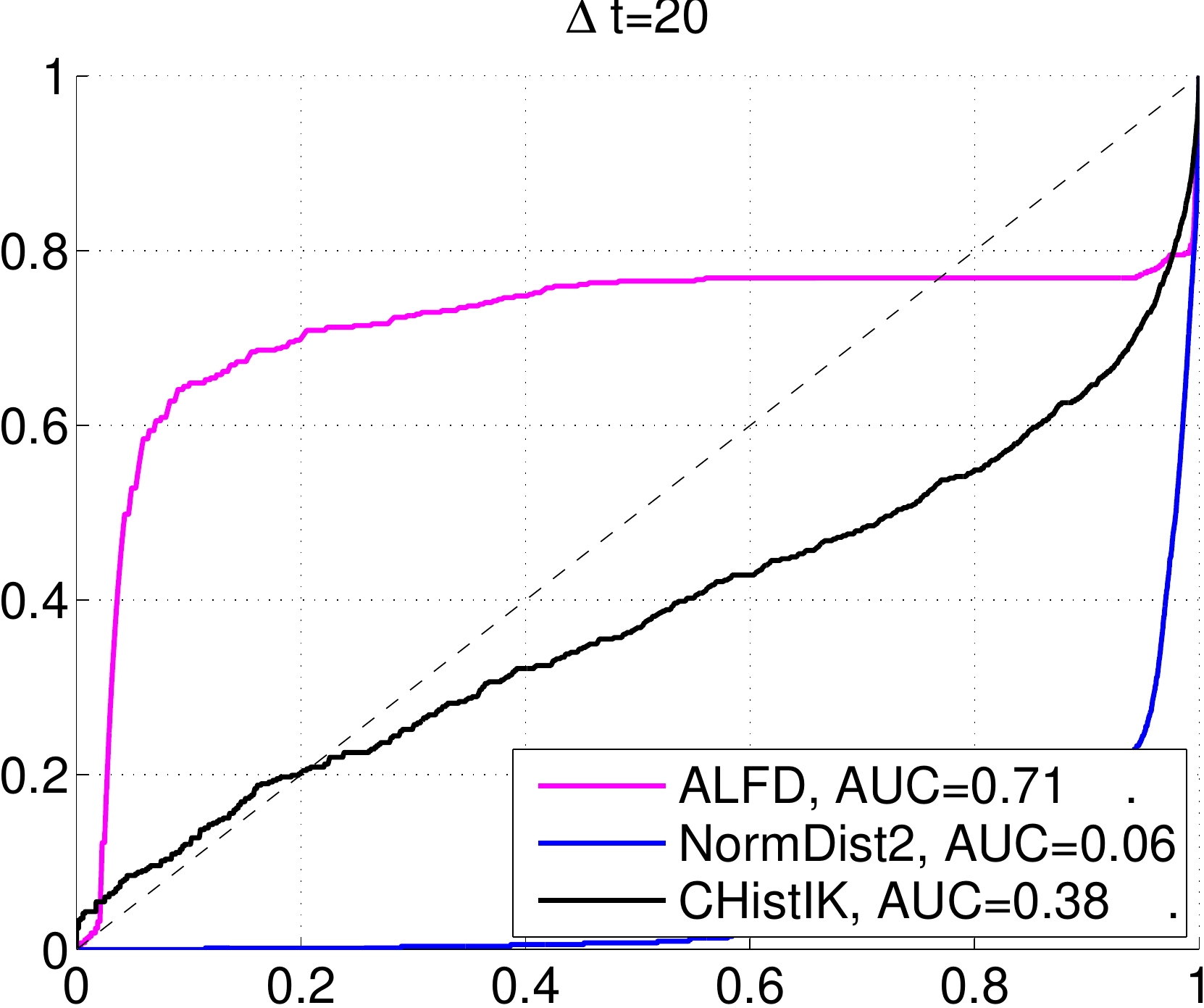} \\
\multicolumn{3}{c}{PETS09-S2L1: Pedestrians, Static camera} \\
\includegraphics[width=0.3\linewidth, trim =22mm 72mm 25mm 66mm, clip]{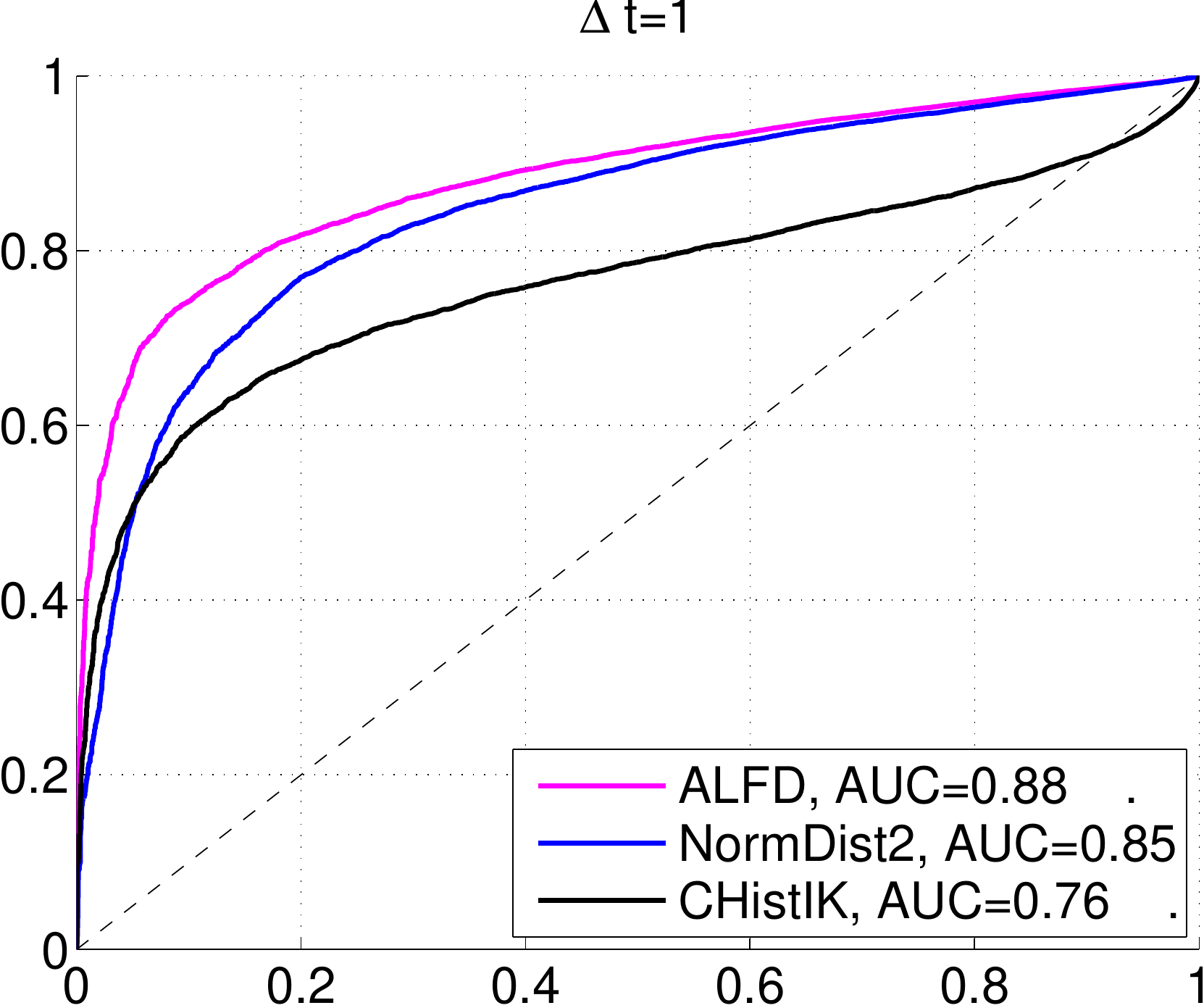} &
\includegraphics[width=0.3\linewidth, trim =22mm 72mm 25mm 66mm, clip]{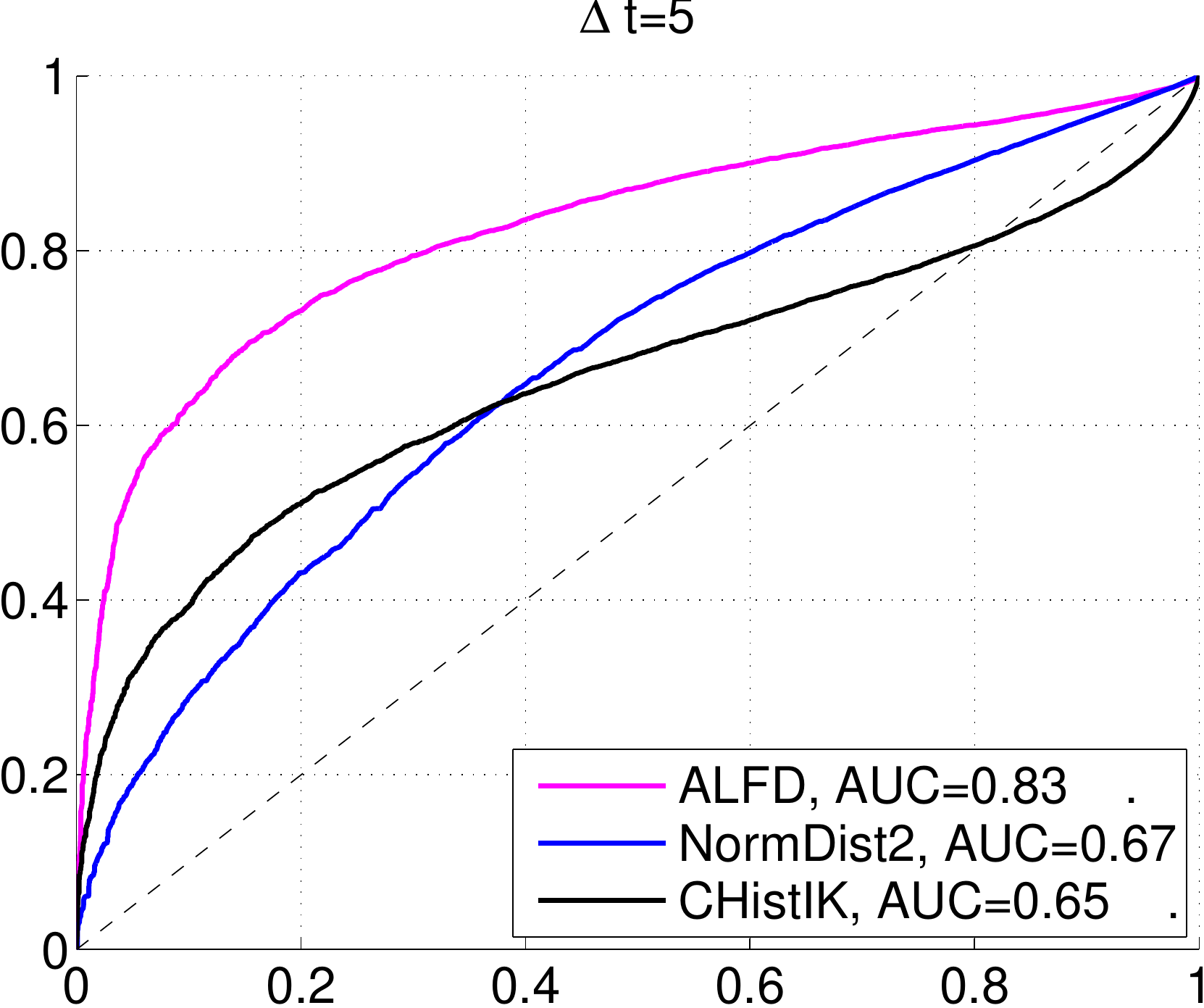} &
\includegraphics[width=0.3\linewidth, trim =22mm 72mm 25mm 66mm,  clip]{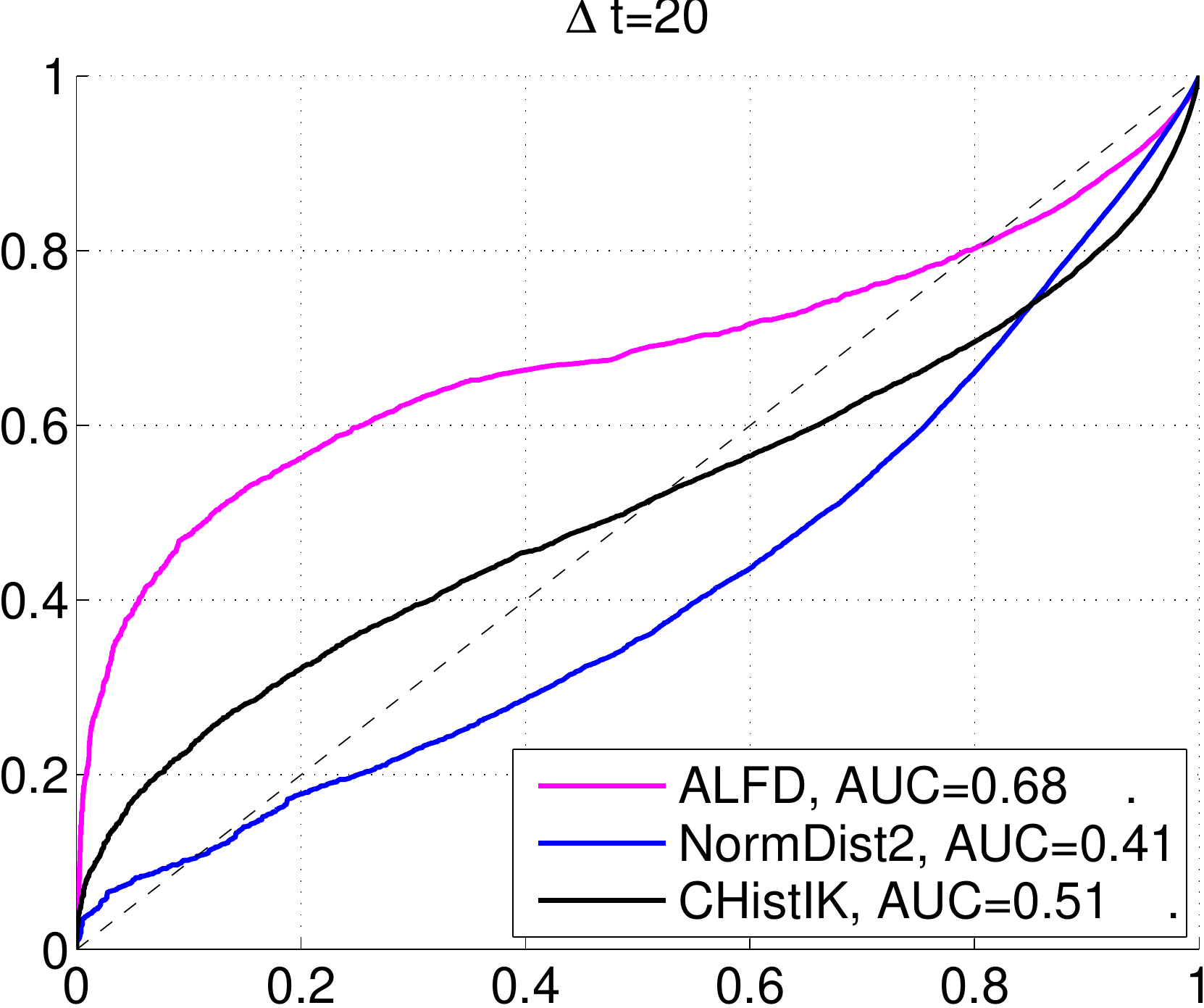} \\
\end{tabular}
}
\end{center}
\caption{Corresponding ROC curves for Table.~\ref{tab:alfd}. $X$ axis is False-Positive-Rate and $Y$ axis is True-Positive-Rate. Notice that NDist2 measure becomes quickly unreliable as the temporal distance increases, when the camera is moving.}
\label{fig:alfdanal}
\end{figure}

We first run an ablative analysis on our ALFD affinity metric. We choose two sequences, KITTI's \emph{0001} and MOT's \emph{PETS09-S2L1} both from the training sets, for the analysis. Given all the detections and the ground truth annotations, we first find the label association between detections and annotations. For each detection, we assign ground truth id if there is larger than $0.5$ overlap. We collect all possible pairs of detections in $1, 2, 5, 10, 20$ frame distance ($\Delta t$), to obtain the positive and negative pairs. As the baseline affinity measures, we use the L2 distance between bottom center of the detections that is normalized by the mean height of the two (NDist2) and the intersection kernel between the color histograms of the two (HistIK). Fig.~\ref{fig:alfdanal} and Table.~\ref{tab:alfd} show the ROC curve and AUC of each affinity metric. We observe that ALFD affinity metric performs the best in all temporal distance regardless of the camera configuration and object type. As the temporal distance increases, the other metrics become quickly unreliable as expected, whereas our ALFD metric still provides strong cue to compare different detections.

\subsection{KITTI Testing Benchmark Evaluation}

Table.~\ref{tab:test} summarizes the evaluation accuracy of our method ({\bf NOMT}) and the other state-of-the-art algorithms on the whole $28$ test video sequences\footnote{The comparison is also available at \url{http://www.cvlibs.net/datasets/kitti/eval_tracking.php} that includes other anonymous submissions.}. 
We also implemented an online tracking algorithm with the Hungarian method~\cite{Kuhn_NRLQ_55} ({\bf HM}) using our unary match function. Any match cost larger than $\minus 0.5$ is set to be an invalid match.
In following evaluations, we set the temporal window $\tau = 10$ and filter out targets that either have only one detection or a median detection score lower than $0$. We use the Kalman Filter~\cite{welch1995introduction} to obtain continuous trajectories out of discrete detection sets $\mathbb{A}$.
Since the KITTI evaluation system does not provide results on \emph{Cyclist} category (due to lack of sufficient data), we report the accuracy of \emph{Car} and \emph{Pedestrian} categories. We also run the experiments with more advanced detection results ({\bf HM+\cite{wang2013regionlets}} and {\bf NOMT+\cite{wang2013regionlets}}).

\begin{table*}
\begin{center}
{\scriptsize
\begin{tabular}{| c || c || c | c | c || c | c | c | c | c | c |}
\hline
& Method & Rec. $\uparrow$ & Prec. $\uparrow$ & F1 $\uparrow$ & MOTA $\uparrow$ & MOTP $\uparrow$ & MT $\uparrow$ & ML $\downarrow$ & IDS $\downarrow$ & FRAG $\downarrow$ \\ 
\hline
\hline
\multicolumn{11}{|c|}{Car Tracking Benchmark} \\
\hline
DPMF~\cite{Pirsiavash_CVPR_11} & Batch & 45.52 \% & {\bf 96.48} \% & 61.85 \% & 43.77 \% & 78.49 \% & 11.08 \% & 39.45 \% & 2738 & 3241 \\
TBD~\cite{Geiger2014PAMI} & Batch & 54.47 \% & 95.44 \% & 69.36 \% & 51.73 \% & 78.47 \% & 13.81 \% & 34.60 \% & 33 & 540 \\
CEM~\cite{Milan:2014:CEM} & Batch & 53.75 \% & 90.31 \% & 67.39 \% & 47.81 \% & 77.26 \% & 14.42 \% & 33.99 \% & 125 & 401 \\
RMOT~\cite{YoonWACV2015} & Online & 55.58 \% & 90.06 \% & 68.74 \% & 49.25 \% & 75.33 \% & 15.17 \% & 33.54 \% & 51 & 389 \\
\hline
HM & Online & 62.13 \% & 94.06 \% & 74.83 \% & 57.55 \% & {\bf 78.79} \% & 26.86 \% & 30.5 \% & 28 & 253 \\
NOMT & Online & {\bf 67.01 } \% & 94.02 \% & {\bf 78.25} \% & {\bf 62.44}  \% & 78.32 \% & {\bf 31.56} \% & {\bf 27.77} \% & {\bf 13} & {\bf 159} \\
\hline
\hline
RMOT~\cite{YoonWACV2015}+\cite{wang2013regionlets} & Online & 78.16 \% & 82.64 \% & 80.34 \% & 60.27 \% & 75.57 \% & 27.01 \% & {\bf 11.38} \% & 216 & 755 \\
\hline
HM+\cite{wang2013regionlets} & Online & 78.47  \% & 90.71 \% & 84.15 \% & 69.12  \% & {\bf 80.10} \% & 38.54 \% & 15.02 \% & 109 & 378\\
NOMT+\cite{wang2013regionlets} & Online & {\bf 80.79}  \% & {\bf 91.00} \% & {\bf 85.59} \% & {\bf 71.68}  \% & 79.55 \% & {\bf 43.10} \% & 13.96 \% & {\bf 39} & {\bf 236}\\
\hline
\hline
\multicolumn{11}{|c|}{Pedestrian Tracking Benchmark} \\
\hline
CEM~\cite{Milan:2014:CEM} & Batch & 46.92 \% & 81.59 \% & 59.58 \% & 36.21 \% & 74.55 \% & 7.95 \% & 53.04 \% & 221 & 1011 \\
RMOT~\cite{YoonWACV2015} & Online & 50.88 \% & 82.51 \% & 62.95 \% & 39.94 \% & 72.86 \% & 10.02 \% & 47.54 \% & 132 & 1081 \\
\hline
HM & Online & 52.28 \% & 83.89 \% & 64.42 \% & 41.67 \% & {\bf 75.77} \% & 11.43 \% & 51.65 \% & 101 & 996 \\
NOMT & Online & {\bf 59.00} \% & {\bf 84.44} \% & {\bf 69.46} \% & {\bf 47.84} \% & 75.01 \% & {\bf 14.54} \% & {\bf 43.10} \% & {\bf 47} & {\bf 959} \\
\hline
\hline
RMOT~\cite{YoonWACV2015}+\cite{wang2013regionlets} & Online & 68.55 \% & 80.76 \% & 74.16 \% & 51.06 \% & 74.19 \% & 16.93 \% & 41.28 \% & 372 & 1515 \\
\hline
HM+\cite{wang2013regionlets} & Online & 67.58 \% & 85.05 \% & 75.32 \% & 54.46  \% & 77.51 \% & 17.31 \% & 42.32 \% & 295 & 1248\\
NOMT+\cite{wang2013regionlets} & Online & {\bf 70.80} \% & {\bf 86.60} \% & {\bf 77.91} \% & {\bf 58.80}  \% & {\bf 77.10} \% & {\bf 23.52} \% & {\bf 34.76} \% & {\bf 102} & {\bf 908}\\
\hline
\end{tabular}
}
\caption{Multiple Target tracking accuracy for KITTI Car/Pedestrian tracking benchmark. $\uparrow$ represents that high numbers are better for the metric and $\downarrow$ means the opposite. The best numbers in each column are bold-faced. We use $\tau = 10$ for NOMT and NOMT+\cite{wang2013regionlets}.}
\label{tab:test}
\end{center}

\end{table*}

\begin{table*}
\begin{center}
{\scriptsize
\begin{tabular}{| c || c || c | c || c | c | c | c | c | c |}
\hline
& Method & FP $\downarrow$ & FN $\downarrow$ & MOTA $\uparrow$ & MOTP $\uparrow$ & MT $\uparrow$ & ML $\downarrow$ & IDS $\downarrow$ & FRAG $\downarrow$ \\ 
\hline
\hline
\multicolumn{10}{|c|}{Pedestrian Tracking Benchmark} \\
\hline
DP~\cite{Pirsiavash_CVPR_11} & Batch & 13,171 & 34,814 & 14.5 \% & 70.8 \% & 6.0 \% & 40.8 \% & 4,537 & 3,090 \\
TBD~\cite{Geiger2014PAMI} & Batch & 14,943 & 34,777 & 15.9 \% & 70.9 \% & 6.4 \%	& 47.9 \% & 1,939 &	1,963 \\	
RMOT~\cite{YoonWACV2015} & Online & 12,473 & 36,835 & 18.6 \% & 69.6 \% & 5.3 \% & 53.3 \% & 684 & 1,282\\
CEM~\cite{Milan:2014:CEM} & Batch & 14,180 & 34,591 & 19.3 \% & 70.7 \% & 8.5 \% & 46.5 \% & 813 & 1,023 \\
\hline
HM & Online & 11,162 & 33,187 & 26.7 \% & 71.5 \% &  11.2 \% & 47.9 \% & 669 & 916 \\
NOMT & Online & {\bf 7,762} & {\bf 32,547} & {\bf 33.7} \% & {\bf 71.9} \% & {\bf 12.2} \% & {\bf 44.0} \% & {\bf 442} & {\bf 823} \\
\hline
\end{tabular}
}
\caption{Multiple Target tracking accuracy for MOT Challenge. $\uparrow$ represents that high numbers are better for the metric and $\downarrow$ means the opposite. The best numbers in each column are bold-faced. We use $\tau = 10$ for NOMT.}
\label{tab:mot}
\end{center}
\end{table*}

As shown in the table, we observe that our algorithm (NOMT) outperforms the other state-of-the-art methods in most of the metrics with significant margins. Our method produces much larger numbers of \emph{mostly tracked targets} (MT) in both \emph{Car} and \emph{Pedestrian} experiments with smaller numbers of \emph{mostly lost targets} (ML). This is thanks to the highly accurate identity maintenance capability of our algorithm demonstrated in the low number of \emph{identity switch} (IDS) and \emph{fragmentation} (FRAG). In turn, our method achieves highest MOTA compared to other state-of-the-arts ($> 10 \%$ for Car and $> 8 \%$ for Pedestrian), which summarize all aspects of tracking evaluation. Notice that the higher tracking accuracy results in the higher detection accuracy as shown in \emph{Recall}, \emph{Precision}, and \emph{F1} metrics. Our own HM baseline also performs better than the other state-of-the-art methods, which demonstrates the robustness of ALFD metric. However, due to the nature of pure online association and lack of high order potential, it ends up missing more targets as shown in the MT and ML measures.  

\subsection{MOT Challenge Evaluation}

Table.~\ref{tab:mot} summarizes the evaluation accuracy of our method ({\bf NOMT}) and the other state-of-the-art algorithms on the MOT test video sequences\footnote{The comparison is also available at \url{http://nyx.ethz.ch/view_results.php?chl=2}.}. The website provides a set of reference detections obtained using \cite{DollarPAMI14pyramids}.

\begin{figure*}
{\footnotesize
\begin{tabular}{@{\hspace{.1mm}}c@{\hspace{.1mm}}c@{\hspace{.1mm}}c@{\hspace{.1mm}}c@{\hspace{.1mm}}}
MOT : AVG-TownCentre @ 237 & MOT : TUD-Crossing @ 70 & MOT : PETS09-S2L1 @ 306 & MOT : PETS09-S2L2 @ 140 \\
\includegraphics[height=0.17\linewidth]{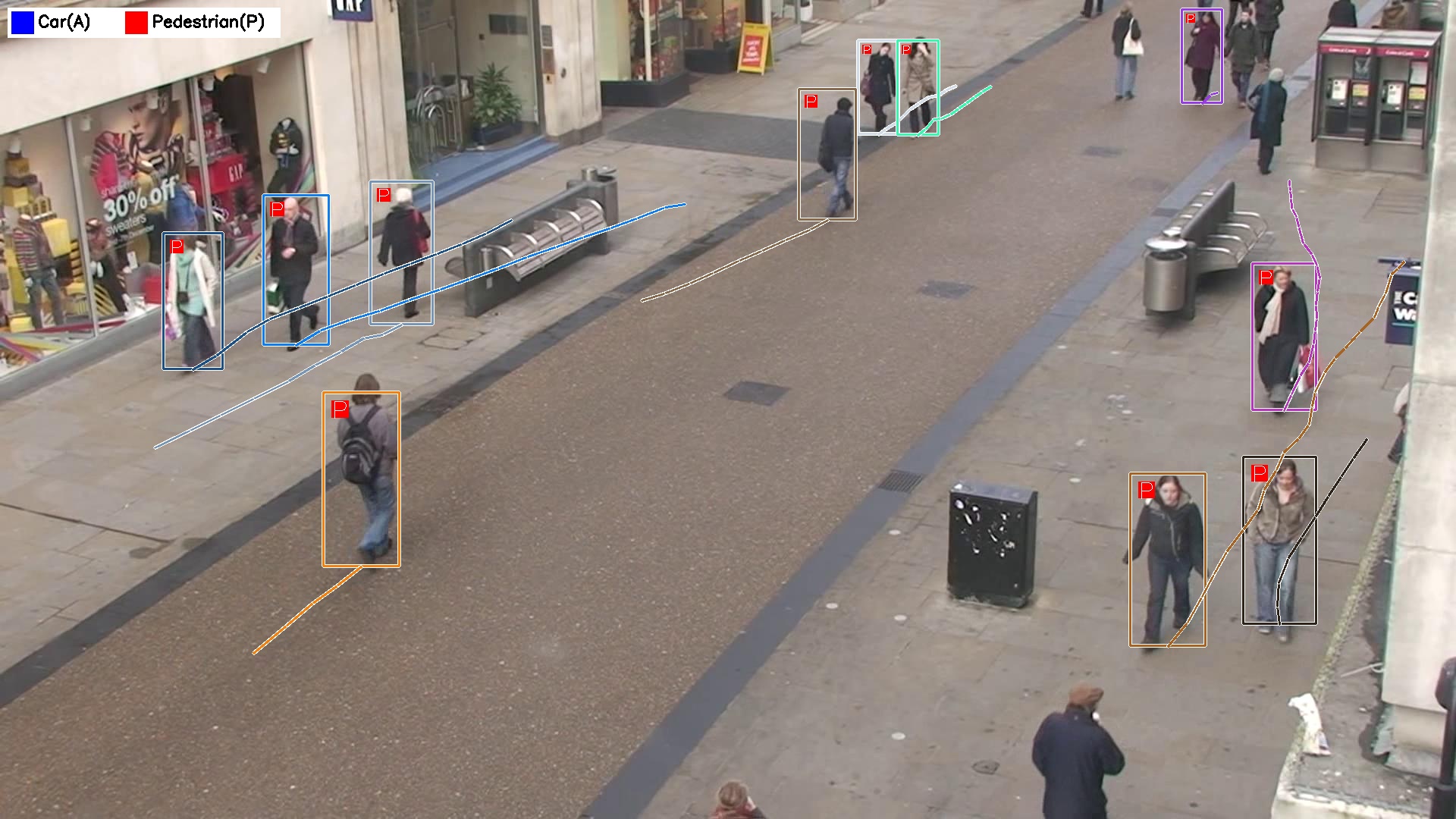}&
\includegraphics[height=0.17\linewidth]{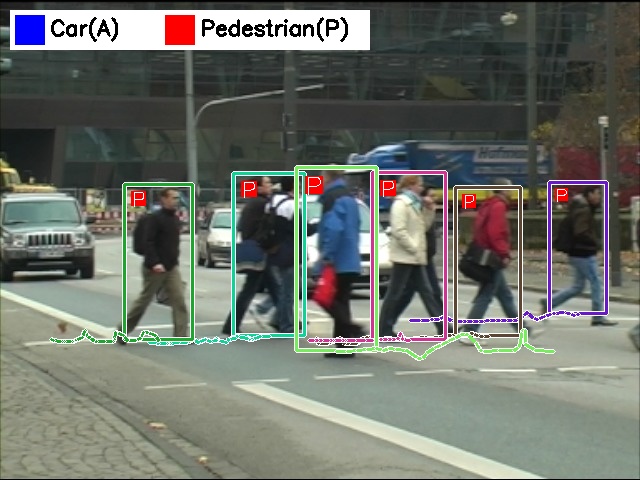}&
\includegraphics[height=0.17\linewidth]{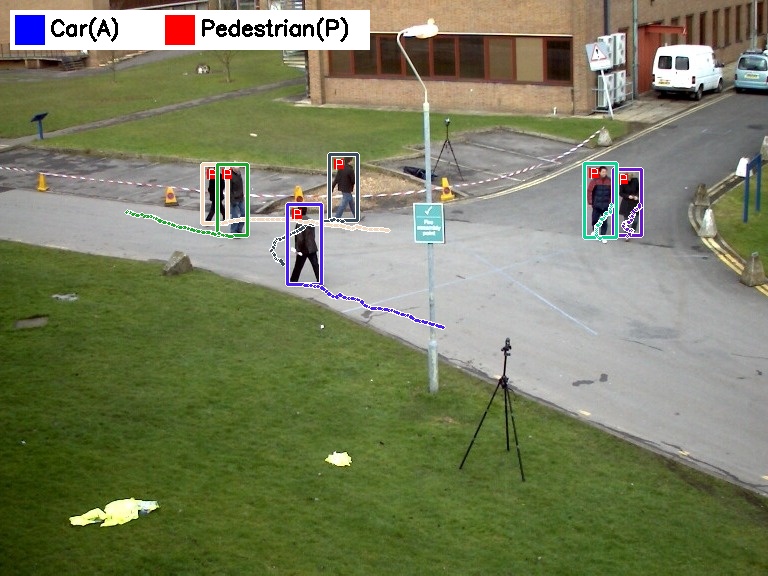}&
\includegraphics[height=0.17\linewidth]{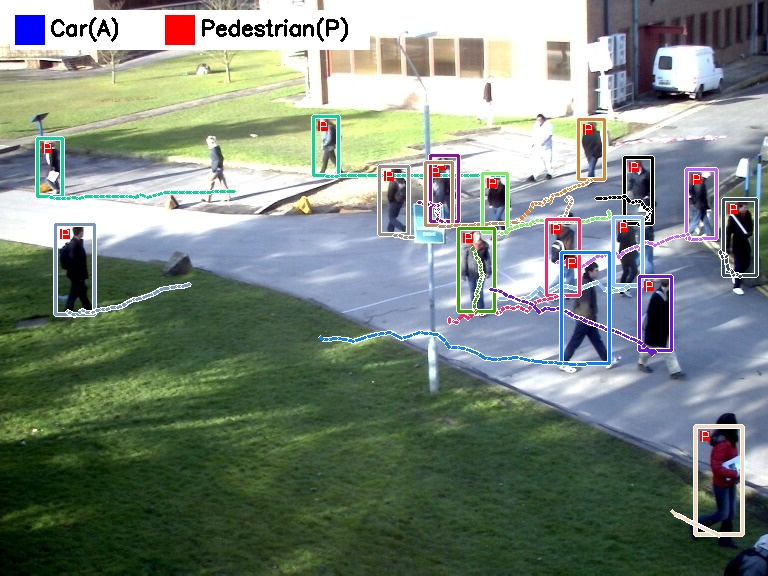}\\
\end{tabular}
\begin{tabular}{@{\hspace{.1mm}}c@{\hspace{.1mm}}c@{\hspace{.1mm}}c@{\hspace{.1mm}}}
KITTI Train 0001 @ 225 & KITTI Train 0009 @ 147 & KITTI Train 0017 @ 34 \\
\includegraphics[width=0.33\linewidth]{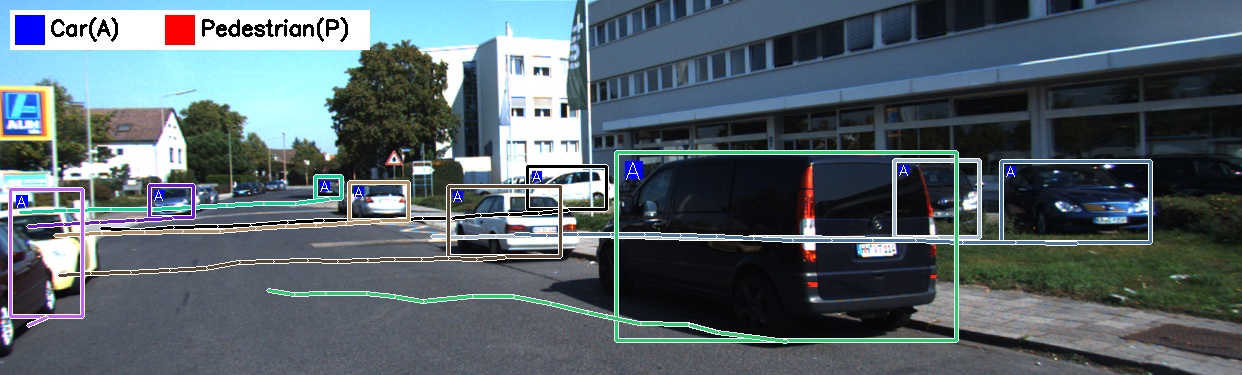}&
\includegraphics[width=0.33\linewidth]{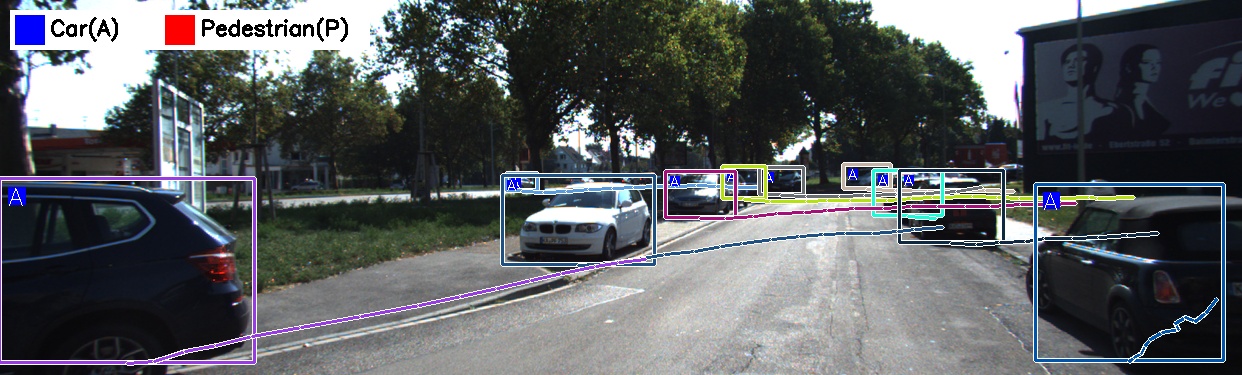}&
\includegraphics[width=0.33\linewidth]{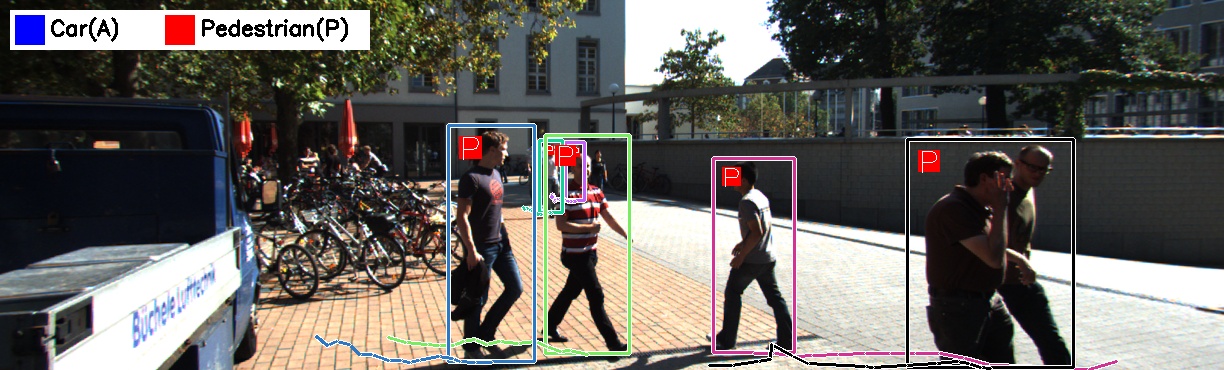}\\	
KITTI Test 0007 @ 78 & KITTI Test 0010 @ 73 & KITTI Test 0016 @ 340 \\
\includegraphics[width=0.33\linewidth]{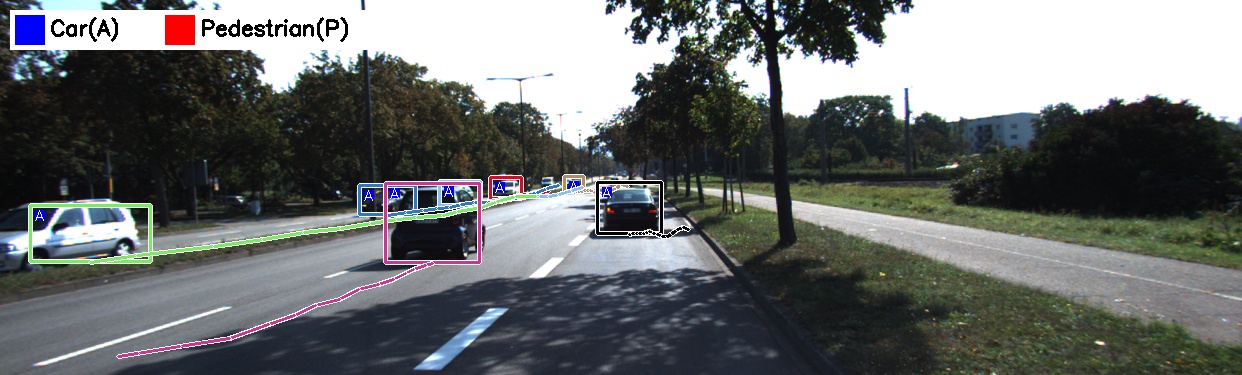}&
\includegraphics[width=0.33\linewidth]{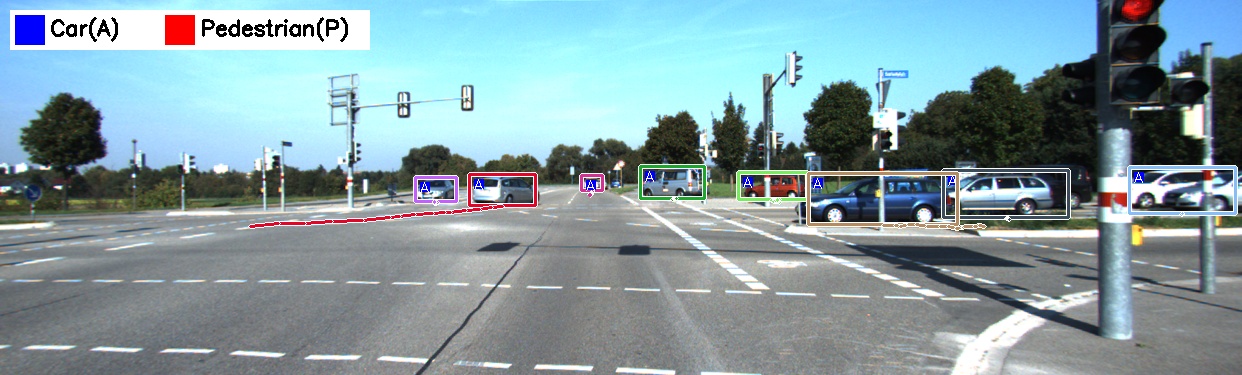}&
\includegraphics[width=0.33\linewidth]{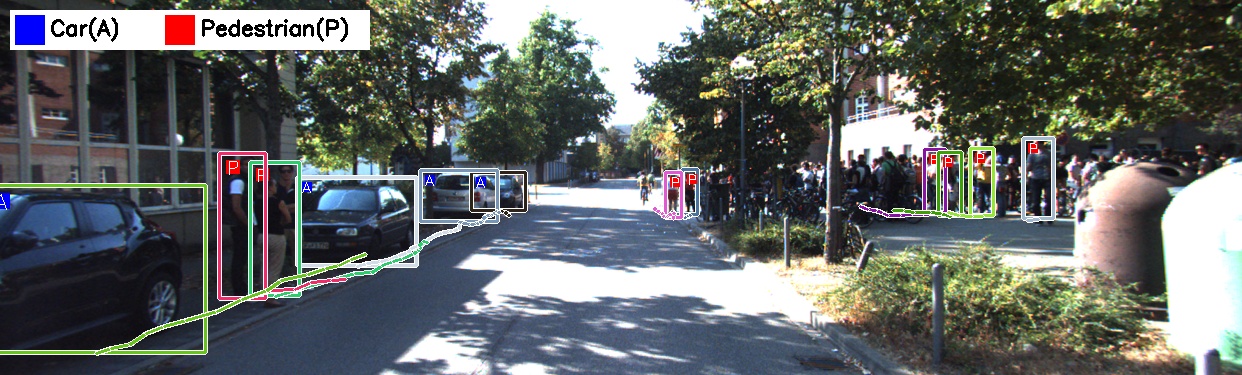}\\	
\end{tabular}}
\caption{Qualitative examples of the tracking results. We show the bounding boxes together with the past trajectories (last $30$ and $10$ frames for MOT and KITTI, respectively). The color of the boxes and trajectories represents the identity of the targets. Notice that our method can generate long trajectories with consistent IDs in challenging situations, such as  occlusion, fast camera motion, etc. The figure is best shown in color.}
\label{fig:qualex}
\end{figure*}

Similarly to the KITTI experiment, we observe that our algorithm outperforms the other state-of-the-art methods with significant margins. Our method achieves the lowest \emph{identity switch} and \emph{fragmentation} while achieving the highest detection accuracy (lowest \emph{False Positives} (FP) and \emph{False Negatives} (FN)). In turn, our method records the highest MOTA compared to the other state-of-the-arts with a significant margin ($> 14 \%$). The two experiments demonstrate that our ALFD metric and NOMT algorithm is generally applicable to any application scenario. Fig.~\ref{fig:qualex} shows some qualitative examples of our results.

\begin{table}[t]
{\scriptsize
\begin{tabular}{|@{\hspace{1.5mm}}c@{\hspace{1.5mm}}||@{\hspace{1.5mm}}c@{\hspace{1.5mm}}||@{\hspace{1.5mm}}c@{\hspace{1.5mm}}|@{\hspace{1.5mm}}c@{\hspace{1.5mm}}|@{\hspace{1.5mm}}c@{\hspace{1.5mm}}|@{\hspace{1.5mm}}c@{\hspace{1.5mm}}||@{\hspace{1.5mm}}c@{\hspace{1.5mm}}|}
\hline
Dataset & FPS & IPT & CHist & Hypos & Infer & Total\\
\hline
KITTI (11,095) & 10.27 & 644.2 & 238.8 & 236.0 & 15.6 & 1,080.2 \\
KITTI+\cite{wang2013regionlets} (11,095) & 10.15 & 615.6 & 161.5 & 144.9 & 40.3 & 1,092.5 \\
\hline
MOT (5,783) & 11.5 & 323.4 & 92.7 & 62.1 & 19.6 & 502.5  \\
\hline
\end{tabular}
}
\caption{Computation time on KITTI and MOT test datasets. The total number of images is shown in parentheses. We report the average FPS (images/total) and the time (seconds) spent in IPT computation (IPT), Color Histogram extraction (CHist), Hypothesis generation (Hypos) that includes all the potential computations, and the CRF inference (Infer). Total time includes file IO (reading images). The main bottleneck is the optical flow computation in IPT module, that can be readily improved using a GPU architecture.}
\label{tab:time} 
\end{table}

\subsection{Timing Analysis}

In order to understand the timeliness of the NOMT method, we measure the latency by computing the difference between detection time ($t_i$ of $d_i$ in $\mathbb{A}^T$) and the last association time. The last association time is defined as: if a detection $d_i$ is newly added to a target $A_m^t$ or replace any other detection $d_j$ (e.g. $t_i = t_j$) in $A_m^{t-1}$ at $t$, $t$ is recorded as the last association time for $d_i$. If $d_i$ was in the $A_m^{t-1}$, no change is made to the last association time of $d_i$. The last association time tells us when the algorithm first recognizes the $d_i$ as a part of $A_m^T$ (the final trajectory output for the target $m$). The mean and standard deviation are $0.59 \pm 1.75$ and $0.66 \pm 1.87$ with \cite{wang2013regionlets} for the KITTI test set ($84.7\%$ and $83.9\%$ with no latency) and $0.87 \pm 2.04$ for the MOT test set ($77.6\%$ with no latency). It shows that NOMT is indeed a near online method.

Our algorithm is not only highly accurate, but also very efficient. Leveraging on the parallel computation, we achieve a real-time efficiency ($\sim 10 FPS$) using a 2.5GHz CPU with 16 cores. Table.~\ref{tab:time} summarizes the time spent in each computational module. 

\section{Conclusion}
\label{sec:conc}
In this paper, we propose a novel \emph{Aggregated Local Flow Descriptor} that enables us to accurately measure the affinity between a pair of detections and a \emph{Near Online Muti-target Tracking} that takes the advantages of both the pure online and global tracking algorithms. Our controlled experiment demonstrates that ALFD based affinity metric is significantly better than other conventional affinity metrics. Equipped with ALFD, our NOMT algorithm generates significantly better tracking results on two challenging large-scaler datasets. In addition, our method runs in real-time that enables us to apply the method in a variety of applications including autonomous driving, real-time surveillance, etc.

{\small
\bibliographystyle{ieee}
\bibliography{egbib}
}


\end{document}